\DocumentMetadata{}
\documentclass[acmsmall]{acmart}

\usepackage{url}
\usepackage{hyperref}

\newcommand\ag[1]{\textcolor{black}{#1}}

\newcommand{\head}[1]{\par\noindent\textbf{#1:}\space}

\usepackage{graphicx}
\usepackage{caption}
\usepackage{subcaption}
\usepackage{multirow}
\usepackage{comment}
\usepackage{adjustbox}
\usepackage{enumitem}

\usepackage{framed}

\newcommand{\vadim}[0]{Vadim Liventsev}
\newcommand{\angie}[0]{Anastasiia Grishina}
\newcommand{\aki}[0]{Aki H\"{a}rm\"{a}}
\newcommand{\leon}[0]{Leon Moonen}

\newcommand{\vadimorcid}[0]{0000-0002-6670-6909}
\newcommand{\angieorcid}[0]{0000-0003-3139-0200}
\newcommand{\akiorcid}[0]{0000-0002-2966-3305}
\newcommand{\leonorcid}[0]{0000-0002-1761-6771}

\newcommand{\method}[0]{SEIDR}
\newcommand{\synthesize}[0]{SYNTHESIZE}
\newcommand{\execute}[0]{EXECUTE}
\newcommand{\instruct}[0]{INSTRUCT}
\newcommand{\instructs}[0]{INSTRUCT$^{\text{static}}$}
\newcommand{\instructllm}[0]{INSTRUCT$^{\text{LLM}}$}
\newcommand{\debug}[0]{DEBUG}
\newcommand{\rank}[0]{RANK}
\newcommand{\beamwidth}[0]{W}
\newcommand{\treearity}[0]{N}
\newcommand{\treearitydebug}[0]{N_\text{debug}}
\newcommand{\treearityexplain}[0]{N_\text{explain}}
\newcommand{\treearitydraft}[0]{N_\text{synth}}
\newcommand{\expectedoutput}[0]{O}

\newcommand{\debugmodel}[0]{$p_\text{debug}(\text{code}, \text{descr})$}
\newcommand{\textmodel}[0]{$p_\text{explain}(\text{code}, \text{descr})$}
\newcommand{\synthmodelnoargs}[0]{$p_\text{synth}$}
\newcommand{\debugmodelnoargs}[0]{$p_\text{debug}$}
\newcommand{\textmodelnoargs}[0]{$p_\text{explain}$}

\newcommand{\gpt}[0]{GPT-3.5}
\newcommand{\llama}[0]{Llama~3}
\newcommand{\cpp}[0]{C++}
\newcommand{\py}[0]{Python}

\newenvironment{DIFnomarkup}{}{}

\newcommand{\rqtreearity}[0]{RQ1}

\newcommand{\rqllama}[0]{RQ2}
\newcommand{\rqmultirun}[0]{RQ3}
\newcommand{\rqlexicase}[0]{RQ4}

\newcommand{\smalltt}[1]{\texttt{\fontsize{8.5}{9}\selectfont#1}}

\AtBeginDocument{%
  }

\setcopyright{cc}
\setcctype{by}
\acmJournal{TELO}
\acmYear{2025} 
\acmVolume{} 
\acmNumber{} 
\acmArticle{} 
\acmMonth{2} 
\acmDOI{10.1145/3719351}

\begin{document}

\title{Fully Autonomous Programming using Iterative Multi-Agent Debugging with Large Language Models}

\author{\angie}
\email{anastasiia@simula.no}
\orcid{\angieorcid}
\authornotemark[1]
\affiliation{%
  \institution{Simula} %
  \city{Oslo}
  \country{Norway}
}
\affiliation{%
  \institution{University of Oslo} %
  \city{Oslo}
  \country{Norway}
}

\author{\vadim}
\email{v.liventsev@tue.nl}
\orcid{\vadimorcid}
\authornote{~Both authors contributed equally to this research.}
\affiliation{%
  \institution{TU Eindhoven}
  \city{Eindhoven}
  \country{The Netherlands}
}
\affiliation{%
  \institution{Philips Research}
  \city{Eindhoven}
  \country{The Netherlands}
}

\author{\aki}
\email{aki.harma@philips.com}
\orcid{\akiorcid}
\affiliation{%
  \institution{Philips Research}
  \city{Eindhoven}
  \country{The Netherlands}}

\author{\leon}
\email{leon.moonen@computer.org} %
\orcid{\leonorcid}
\authornote{~Corresponding author.}
\affiliation{%
  \institution{Simula}
  \city{Oslo}
  \country{Norway}%
}

\renewcommand{\shortauthors}{Grishina and Liventsev et.al.}

\begin{abstract}

Program synthesis with Large Language Models (LLMs) suffers from a ``near-miss syndrome'': the generated code closely resembles a correct solution but fails unit tests due to minor errors. We address this with a multi-agent framework called Synthesize, Execute, Instruct, Debug, and Repair (SEIDR). Effectively applying SEIDR to instruction-tuned LLMs requires determining (a) optimal prompts for LLMs, (b) what ranking algorithm selects the best programs in debugging rounds, and (c) balancing the repair of unsuccessful programs with the generation of new ones. We empirically explore these trade-offs by comparing replace-focused, repair-focused, and hybrid debug strategies. 
We also evaluate lexicase and tournament selection to rank candidates in each generation. On Program Synthesis Benchmark 2 (PSB2), our framework outperforms both conventional use of OpenAI Codex without a repair phase and traditional genetic programming approaches. SEIDR outperforms the use of an LLM alone, solving 18 problems in C++ and 20 in Python on PSB2 at least once across experiments. To assess generalizability, we employ GPT-3.5 and Llama 3 on the PSB2 and HumanEval-X benchmarks. Although SEIDR with these models does not surpass current state-of-the-art methods on the Python benchmarks, the results on HumanEval-C++ are promising. SEIDR with Llama 3-8B achieves an average pass@100 of 84.2\%. Across all SEIDR runs, 163 of 164 problems are solved at least once with GPT-3.5 in HumanEval-C++, and 162 of 164 with the smaller Llama 3-8B. We conclude that SEIDR effectively overcomes the near-miss syndrome in program synthesis with LLMs.

\end{abstract}



\received[accepted]{19 February 2025}

\makeatletter
\fancypagestyle{standardpagestyle}{%
  \fancyfoot[RO,LE]{\footnotesize Accepted for publication in \@journalNameShort, \@acmPubDate.}%
}
\fancypagestyle{firstpagestyle}{%
  \fancyhead[RO,LO]{}
  \fancyfoot[RO]{\footnotesize Accepted for publication in \@journalNameShort, \@acmPubDate.}%
}  
\makeatother

\maketitle
\pagestyle{standardpagestyle}

\section{Introduction}
\label{sec:intro}

Automatic programming has been an important goal in the field of artificial intelligence almost since its inception, promising to reduce the workload of software developers by automatically solving some of the tasks they face~\cite{manna1971:automatic}.
More recently, program synthesis has emerged as an interpretable alternative to black-box machine learning methods that lets human experts understand, validate, and edit the algorithms generated by artificial intelligence~\cite{bastani2022:interpretable}.
In addition to the scientific benefits of such knowledge, it extends the benefits of machine learning to domains, such as embedded systems, where it is technically challenging~\cite{dhar2021:survey}, or healthcare, where it is avoided for safety reasons~\cite{connolly2023:systematic,jia2022:role}.

The predominant methodology in automatic programming has shifted from deductive programming~\cite{manna1992:fundamentals,alur2015:syntaxguided} to genetic and evolutionary methods~\cite{ahvanooey2019:survey} to, more recently, large autoregressive language models trained on corpora of source code~\cite{xu2022:systematic, zan2023:large} to benefit from their remarkable capability for zero-shot generalization~\cite{chen2021:evaluating}.
However, even state-of-the-art models fine-tuned on a specific class of programming tasks still require costly filtering to discard the outputs that fail to compile or pass tests \cite{li2022:competitionlevel}.
These outputs tend to be superficially similar to the correct solutions~\cite{ren2020:codebleu,liu2023:your, shirafuji2023:exploring} despite failing to produce the expected output, an issue known as ``near-miss syndrome'' or ``last mile problem''~\cite{bavishi2022:neurosymbolic}. 

Given these challenges, research in machine learning on source code~\cite{allamanis2018:survey} tends to focus on restricted domain-specific languages~\cite{chen2021:latent,polozov2015:flashmeta,liventsev2021:bf} or automating specific parts\footnote{~Similarly to autonomous driving~\cite{grigorescu2020:survey,marcano2020:review}.} of the software development process~\cite{lu2021:codexglue,niu2023:crosscodebench}, such as code search~\cite{di2023code, kim2023big}, code translation~\cite{yan2023codetransocean}, detection of issues~\cite{fernandes2016:reviewbased,chakraborty2021:deep}, improvement~\cite{petke2018:genetic} and repair~\cite{legoues2019:automated} rather than fully autonomous programming in a programming language popular with human developers.\footnote{~\url{https://www.tiobe.com/tiobe-index/}}
However, two recent innovations potentially make the latter task tractable.

One is \emph{Synthesize, Execute, Debug}~\cite{gupta2020:synthesize}, a framework that attempts to bridge the ``last mile'' gap by introducing program repair into the program synthesis algorithm. 
A programming task is specified using both a natural language description and a set of input/output (I/O) pairs that demonstrate what output is expected from the program, thereby combining text-to-code~\cite{iyer2018:mapping} and programming by example~\cite{halbert1984:programming,gulwani2016:programming} paradigms typical for competitive programming~\cite{zavershynskyi2018:naps}.
\emph{Synthesize, Execute, Debug} creates a first draft program using a generative model, compiles and executes it with given input examples.
This is followed by a program repair step to fix the identified errors.

Another relevant innovation is instruction fine-tuned large language models~\cite{zhang2024:instruction}. Instruction fine-tuned models use human feedback in their training process and are designed to explicitly or implicitly admit two inputs: a source text (or code) and a textual command instructing the model to edit the source in a particular way, e.g., ``summarize'' or ``translate to Python.''
These models have been shown to be highly successful in automatic program repair~\cite{fan2023:automated}. 
However, given the free-form nature of these instructions,\footnote{~\ag{Throughout this paper we avoid other definitions of \emph{instruction}, such as \emph{an individual operation in code}, to prevent ambiguity.}} how one should engineer instructions that maximize repair performance is an open question. 

These innovations have led to the proposal of a framework, \emph{Synthesize, Execute, Instruct, Debug and Rank}, or \method{},\footnote{~Seiðr also refers to a type of Norse magic~\cite{blain2002:nine} pertaining to predicting and controlling the future, which we deem thematically appropriate.} initially introduced in our GECCO-2023 paper~\cite{liventsev2023:fully}. 
SEIDR is a multi-agent iterative framework that uses feedback from code execution and failing test cases to update the initially generated buggy code. 
Our initial explorations cover SEIDR with GPT-3 as the bug summarization model and Codex (or GPT-3 trained on code) as the program generation and debugging model.  
In addition, by modifying the tree arity parameter (see Section~\ref{sec:beam-search}), we investigate the trade-off between generating and repairing only one program versus regenerating any program that does not pass all the tests, as well as intermediate configurations, where we build a tree of programs and update the best ones.

While Codex is an early code generation model, the emergence of new models that score better in programming and natural languages motivates further research into the use of \method{} with newer models. 
To study the generalizability of the initial SEIDR results, we use two other LLMs, Llama 3\footnote{~\url{https://ai.meta.com/blog/meta-llama-3/}} and GPT-3.5,\footnote{~\url{https://openai.com/form/researcher-access-program/}} and an additional dataset, HumanEval-X, with different tree arity parameters~\cite{brown2020:language, chen2021:evaluating, zheng2023:codegeex}. 
Moreover, we build up on the initial experiments with Codex and zoom in on the area with the best-performing tree arities in a hyperparameter search for a better repair-replace trade-off resolution. 

To reflect on the parent selection strategies used in the Rank agent of \method{}, we also explore whether the programs should be chosen based on the average performance across all tests or whether \method{} can benefit from keeping such programs in the loop that fully cover individual tests, but do not perform well on average.
Therefore, as an alternative to the tournament selection, we test the best tree arity setups with lexicase selection-based ranking~\cite{helmuth2015:solving}.
Moreover, since language models bring in stochasticity, we run the experiments several times to measure the variability of results obtained with fixed hyperparameters and reflect on the results repeatability.

Overall, the current paper contributes to the field of Large Language Models for Software Engineering (LLM4SE) with a framework for code generation and its iterative repair. 
Compared to the earlier GECCO-2023 paper, this article refines and extends the preliminary results by providing a more in-depth explanation of SEIDR as a multi-agent framework for autonomous programming, extending the original experiments and analysis to include HumanEval-X as an additional benchmark, evaluating GPT-3.5 and Llama 3 as additional LLMs, investigating lexicase selection as an alternative to tournament selection, and conducting a repeatability analysis of the framework’s performance over multiple independent runs. Overall, this extension adds three research questions and doubles the number of configurations considered in these questions compared to our preliminary work. 

Section~\ref{sec:methodology} presents a framework that adapts \emph{Synthesize, Execute, Debug} to instruction-tuned Large Language Models in agents that can solve programming tasks in an autonomous fashion. 
We discuss related work in Section~\ref{sec:related-work}, introduce experiments to explore effective search 
strategies for this framework in Section~\ref{sec:eval}. 
Finally, in Section~\ref{sec:results}, we demonstrate that our framework outperforms conventional automatic programming techniques, such as genetic programming and naive application of large language models that generate one solution per problem without updating it iteratively. 

\section{Methodology}
\label{sec:methodology}
\begin{figure*}[t]
    \centering
    \includegraphics[width=\linewidth,trim={0mm 0mm 0mm 0mm}]{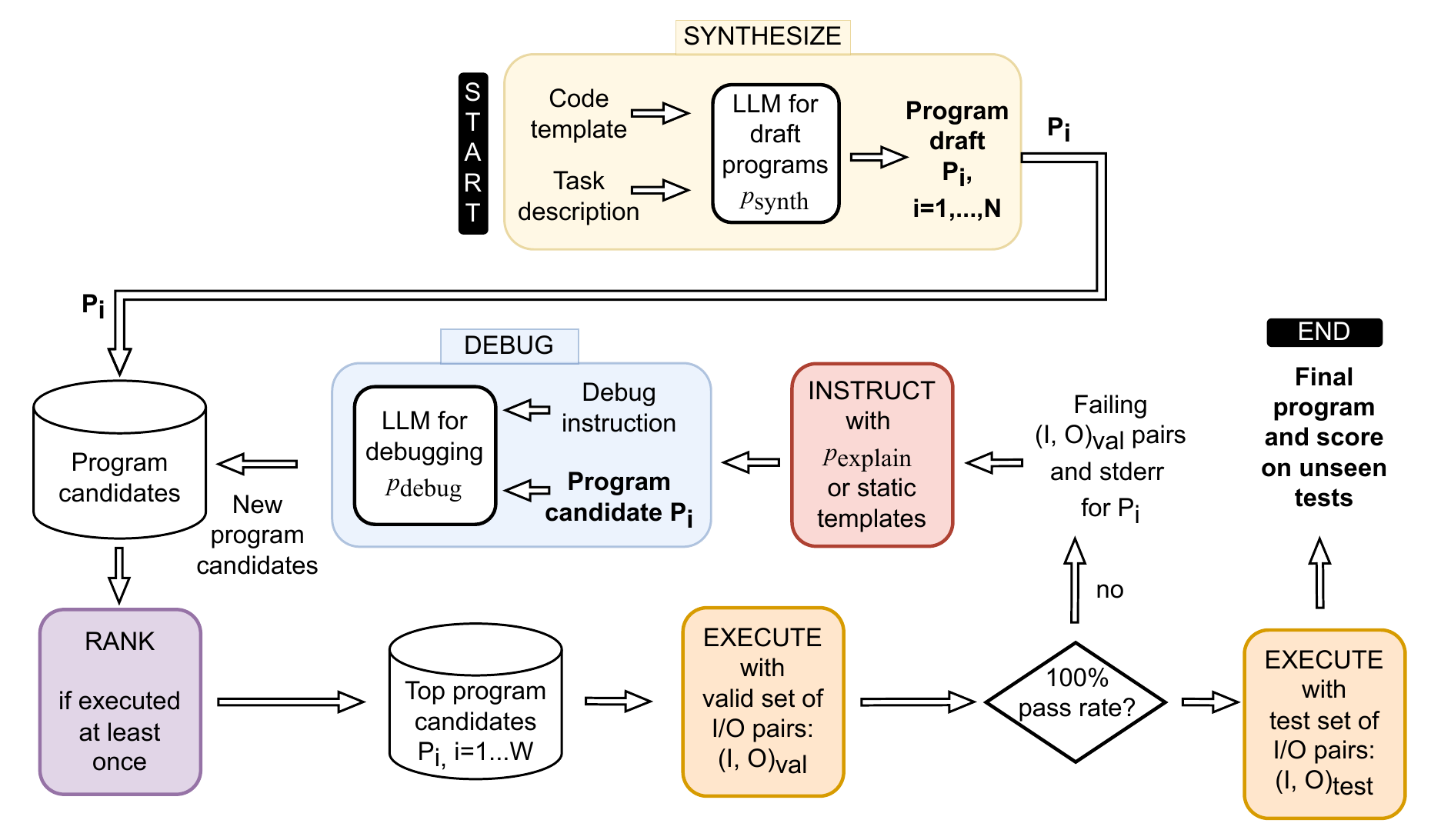}
    \caption{Overview of \method{}, a multi-agent iterative framework that uses LLMs to implement the Synthesize, Execute, Instruct, Debug, and Rank feedback loop.}
    \label{fig:method}
    \Description{SEIDR framework for iterative program synthesis and refinement using LLMs.
The figure illustrates SEIDR, a multi-agent iterative framework that leverages large language models (LLMs) for program synthesis, execution, instruction, debugging, and ranking. The process begins with synthesizing program drafts based on a code template and task description. These drafts form program candidates, which enter an iterative loop. Candidates are executed on validation input-output pairs, and failing programs receive debugging instructions from an LLM or are modified using instructional explanations. The refined candidates are ranked based on execution success. If a program reaches a 100\% pass rate on validation tests, it is evaluated on unseen test input-output pairs, producing a final program and performance score. The workflow is structured as a feedback loop to iteratively improve program correctness.}
\end{figure*}

The proposed five-agent \method{} framework is summarized in Figure~\ref{fig:method}, which we discuss in detail in Section~\ref{sec:ingredients}.
In essence, to solve a programming task defined as a text description and a collection of I/O examples, we split I/O examples into prompt and validation sets and use the prompt set in a large language model to \synthesize{} a population of candidate solutions.
We \execute{} the solutions, test them against the validation set, generate a text description of the identified problems used to \instruct{} a large language model to produce repaired candidate solutions similar to the way a human developer \debug{}s a program.
We \rank{} the candidates
by correctness, measured as matching I/O pairs, discard the worst candidates, and repeat until a fully correct solution is found.

\subsection{Ingredients}
\label{sec:ingredients}

\method{} makes use of instruction fine-tuned large language models: a \emph{synthesis} model $p_{\text{synth}}(\text{code, }$ descr), a \emph{debugging} model \debugmodel{}, as well as a model \textmodel{} that can be used for writing textual instructions, which are forwarded to the code generation model \debugmodelnoargs{} for code updates. 
Therefore, the design can be described as two agents communicating with each other, whereby one generates code and another provides critical or supervising comments on what should be changed in the generated code. 

The models \synthmodelnoargs{}, \debugmodelnoargs{}, and \textmodelnoargs{} can be either separate or the same model.
The prerequisites are that \synthmodelnoargs{} and \debugmodelnoargs{} models are able to ``understand'' natural language (descr) and partial or full programs (code) and generate code based on them. 
The model \textmodelnoargs{}{} should be able to ``understand'' code and natural language and either autocomplete or generate the debugging instruction from scratch. 
Note that \textmodelnoargs{} is optional, since alternatively the debugging instructions can be generated from failing tests using static pre-defined templates.
In general, \method{} requires sequence-to-sequence generative models for these agents. 
In our experiments, we select models for \synthmodelnoargs{}, \debugmodelnoargs{}, and \textmodelnoargs{} based on the state-of-the-art transformer architectures~\cite{vaswani2017:attention} 
(see Section~\ref{sec:models}). 

Each LLM is a highly parameterised probability distribution over the space of (code, description)-tuples with parameters estimated on a large diverse (i.e., non-task-specific) corpus.
This stochastic nature of language models is an important prerequisite for \method{}, since it allows us to sample batches of diverse candidate solutions from \synthmodelnoargs{}, \debugmodelnoargs{}, and \textmodelnoargs{}. 
We denote the number of outputs generated with $\treearitydraft{},$ $\treearitydebug{},$ and $\treearityexplain{},$ correspondingly.
Moreover, each model generates the most probable and less probable outputs in each batch, which helps diversify problem solving attempts. 
In the following implementation-related subsections, we explain how we vary the number of candidate solutions, debug instructions, and repairs generated in a batch by each LLM in \method{}.

While \method{} is described here as a multi-agent system, it can equally be seen as a form of evolutionary algorithm or genetic programming, where the initialization and mutation steps of the system are performed by LLMs, \synthmodelnoargs{} and \debugmodelnoargs{}, correspondingly.
Throughout this work, we use agent-oriented programming terminology~\cite{shoham1993:agentoriented} and evolutionary optimization terminology interchangeably to try to bridge the gaps between these domains.

\subsubsection{Synthesize}
\label{sec:synth}

\begin{figure}[t]
    \centering
    \includegraphics[width=0.8\linewidth, trim={0mm 40mm 0mm 0mm}, clip]{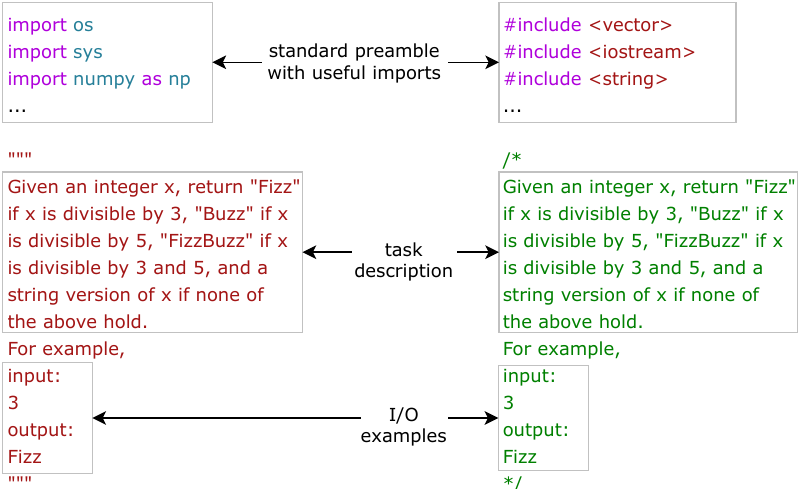}
    \caption{Anatomy of \synthesize{} templates}
    \label{fig:template}
    \Description{Structure of SYNTHESIZE templates for different programming languages.
The figure presents the components of SYNTHESIZE templates, which define code generation structures across programming languages. It includes a standard preamble with necessary imports, shown in Python and C++. The task description, written as comments in both languages, specifies a function that returns “Fizz” if an input integer is divisible by 3, “Buzz” if divisible by 5, “FizzBuzz” if divisible by both, or the number itself otherwise. Input-output examples illustrate expected behavior. The figure highlights how the same problem is expressed as templates in Python and C++.}
\end{figure}

The framework starts with the \synthesize{} agent, which is responsible for generating initial draft solutions.
We start with a basic template for a chosen programming language that contains a number of standard library imports, as shown in Figure~\ref{fig:template}.

We populate this template with a comment indicating a textual task description and several I/O examples from the training set.
We design the templates with prompt engineering guidelines\footnote{~\url{https://platform.openai.com/docs/guides/prompt-engineering/six-strategies-for-getting-better-results}} and prior work~\cite{debruin2021:autoencoders} in mind.
We then sample $\treearitydraft{}$ programs from \synthmodelnoargs{}, setting \texttt{code} to the populated template and \texttt{description} to the natural language description of what the model should generate.
We use spring sampling:\footnote{~\url{https://vadim.me/publications/spring/}} a temperature-based sampling with a monotonically increasing temperature schedule where the $i$-th program is sampled with temperature $t_i \approx \frac{i-1}{\treearitydraft{}}$ (we use approximate equality to enable efficient implementation by means of batching).
Thus, the sampling procedure for the first programs approximates a deterministic maximum-likelihood estimation.
In combination with the naturalness principle of source code \cite{allamanis2018:survey,jiang2022:bugs}, this approach ensures that the samples are diverse, but always contain the most likely programs for the given task.

\subsubsection{Execute}
\label{sec:execute}

The \execute{} agent compiles the programs (if necessary) and launches them using the standard tools for the programming language.
The program is run once for every I/O pair in the validation set. 
Its \texttt{stdin} stream receives all input lines in a given input pair, and its \texttt{stdout} and \texttt{stderr} streams are captured and saved.
We then measure the \emph{score} of the program defined as the accuracy over the output lines, with \expectedoutput{} being the expected output, and $n=\max\{|\expectedoutput{}|, |\text{stdout}|\}$:
\[    
\text{score}(\expectedoutput{}, \text{stdout}) = \frac{\sum^{n}_i{\mathbb{I}[\text{stdout}_i = O_i]}}{n} 
\]
unless \texttt{stderr} is non-empty during compilation or execution, which is considered to indicate failure and is assigned a score of 0.

\subsubsection{Instruct}
\label{sec:instruct}

The goal of the \instruct{} agent is to provide instructions that summarize bugs in a candidate program and suggest a solution for \debugmodelnoargs{}. 
The resulting instructions with the bug summary should indicate what requirement is violated and instruct the LLM to edit the candidate program accordingly. 
The input to \instruct{} consists of failing I/O pairs from the validation set and \texttt{stderr} output of the candidate execution. 
In order to represent this heterogeneous input as text that can be further processed by an LLM, we use template engines that replace placeholders in files or strings with input values and return a formatted string. 
In recent chat- and instruction-based LLMs, the terms \emph{template engine} and \emph{prompt template} are used interchangeably.

\begin{figure*}[t]
    \centering
    \includegraphics[width=0.85\linewidth,trim={0mm 0mm 0mm 0mm}]{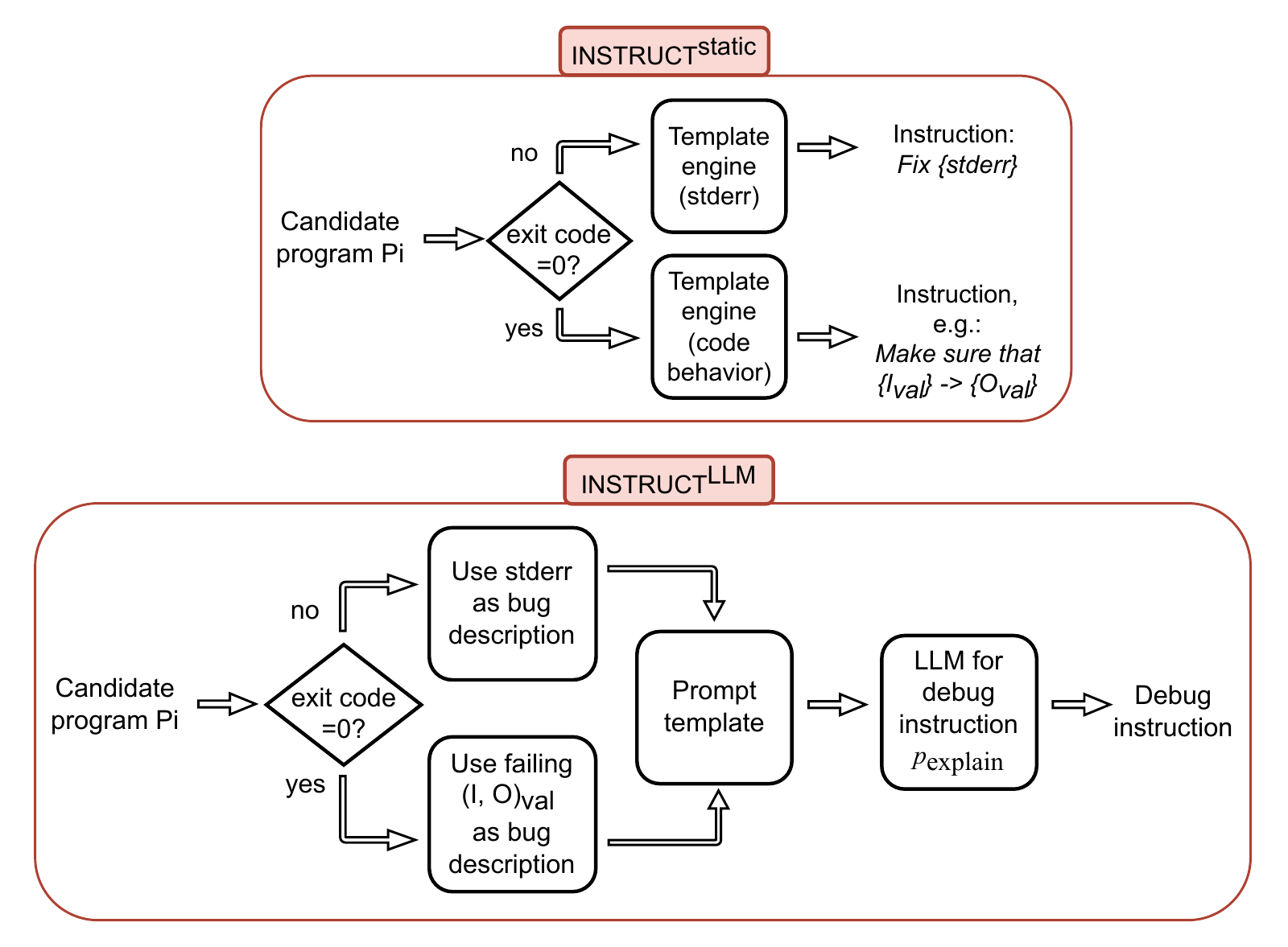}
    \caption{Overview of the two designs for the \instruct{} agent.}
    \label{fig:method-instruct}
    \Description{Comparison of static and LLM-based INSTRUCT agent designs.
The figure illustrates two approaches for generating debugging instructions in the INSTRUCT agent: a static template-based method and an LLM-based method. In both designs, a candidate program is evaluated based on its exit code. If the program fails (nonzero exit code), the static approach uses predefined templates to generate instructions based on error messages (stderr) or incorrect output mappings. In contrast, the LLM-based approach converts stderr and failing input-output pairs into a structured bug description, which is then used in a prompt template for generating debugging instructions via an LLM. The LLM-based method provides more flexible and context-aware guidance compared to the static approach.}
\end{figure*}

We consider two different designs of the \instruct{} agent: \instructs{} and \instructllm{} shown in Figure~\ref{fig:method-instruct}. 
In both cases, if \texttt{stderr} is not empty, i.e., execution exits with code 0 before getting any output to compare it with the expected output, the \texttt{stderr}-based template engine generates the instruction to fix the error. 
However, the designs differ in the way they transform failing I/O pairs to generate instructions in case \texttt{stderr} is empty.

\instructs{} uses a fixed template and substitutes placeholders for input and output with the corresponding strings of the first failing test case in its template engine.
For example, we show the resulting instruction for an exemplar template in Figure~\ref{fig:method-instruct}.
In contrast, \instructllm{} uses the failing I/O pair in the LLM for text completion, thereby prompting the text LLM to produce the bug explanation and a summary for debugging. 
In addition to providing a failing test case or \texttt{stderr}, one may choose to give the model \textmodelnoargs{} more context, such as the problem name, task description, and the code generated so far. 
Each call to \textmodelnoargs{} can result in $\treearityexplain{}\ge1$ instructions, a batch output of an LLM.
The prompt templates used for the experiments are detailed in Section~\ref{sec:prompts}.

\subsubsection{Debug}

The main component of \method{} that addresses the ``near-miss syndrome'' is the \debug{} agent.  
This agent iterates over all programs in the population to repair candidate programs and pass more tests. 
It uses the instructions written by \instruct{} to sample from the \debugmodelnoargs{} model $\treearitydebug{}$ times
to repair each candidate and create a new population of \treearity{} candidates.
For \debugmodelnoargs{}, the parameter \texttt{code} is set to the current version of the candidate solution and \texttt{descr} to the output of \instruct{} and any additional context chosen for a specific implementation.
The current generation of candidates is then replaced by \treearity{} outputs of \debug{}.

\subsubsection{Rank}

The \rank{} agent implements what is known in genetic programming as \emph{parent selection}~\cite{koza1994:genetic}: it selects the best $\beamwidth{}$ programs to be further improved by the \debug{} agent.
We consider two different parent selection algorithms: tournament selection and lexicase selection. 
See Section~\ref{sec:lexicase-results} for their empirical comparison.

\emph{Tournament selection} variant used in this work sorts the programs according to their average test score and selects top $\beamwidth{}$ candidates. 
The programs are selected based on the intuition that repair of the best so far (yet imperfect) programs begets good programs. 
The simple ranking is also referred to as \emph{tournament selection}, where the best-performing candidates are chosen to participate in the next round.  
In the classical tournament selection, the first step is to draw $n$ random candidates from the population. 
In our implementation, the first step is to generate exactly $n$ candidates that are needed for the next round. 
This approach prioritizes candidates that perform the best on average over all tests.

\emph{Lexicase selection}~\cite{helmuth2015:solving} is a ranking approach that maximizes the diversity of selected candidates in addition to their metric-based score.
Lexicase selection ensures diversity by keeping the program candidates that perform the best on unique tests as opposed to the program candidates that perform best on average over all tests.
The algorithm is as follows:
\begin{enumerate}

\setlength{\parskip}{0pt}
\setlength\itemsep{0pt}

    \item randomly shuffle the set of tests;
    \item select a program with the best score on test 1;
    \item if several programs are tied, resolve the tie by selecting the best program on test 2;
    \item repeat for tests $3,4,\dots,$ until only one program is left;
    \item mark this program as ``selected'';
    \item if less than $\beamwidth$ programs are selected, go back to step 1.
\end{enumerate}
This ensures that even if the average quality of the selected candidates is lower, the batch of $\beamwidth{}$ programs collectively contains a higher number of required ``skills'', as measured by tests.

\subsection{Meaning of Hyperparameters}
\label{sec:beam-search}

\begin{figure}[t]
    \centering
    \includegraphics[width=0.7\linewidth, trim={0mm 4mm 0mm 0mm}]{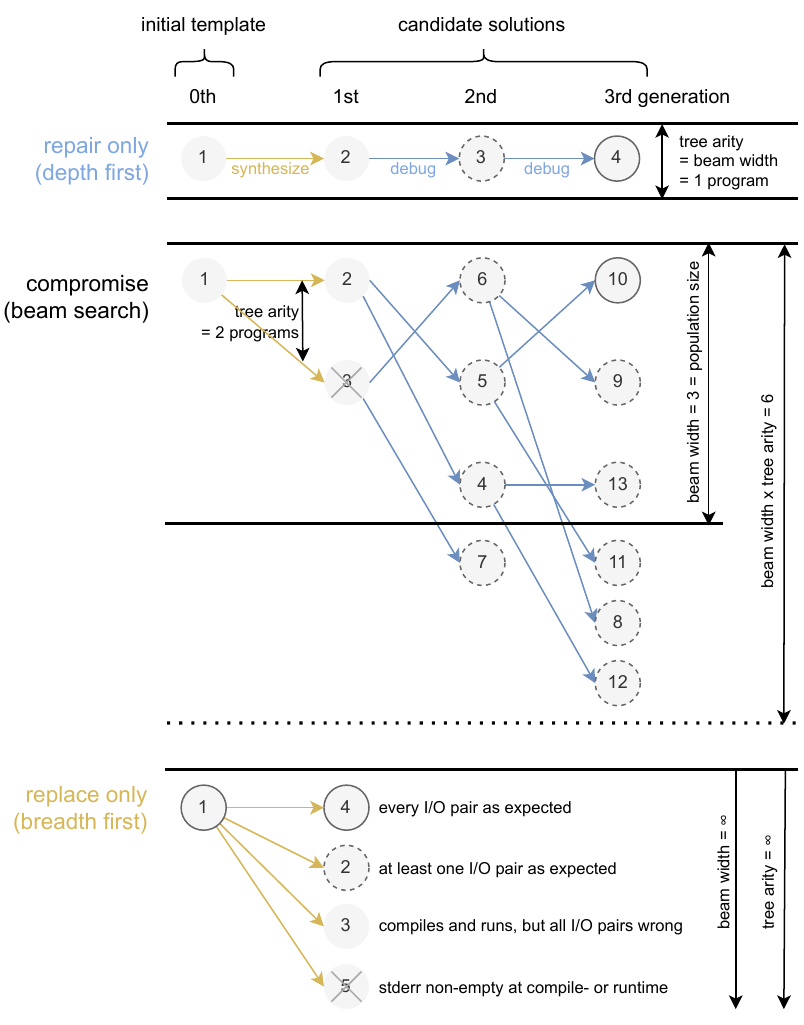}
    \caption{Repair-replace trade-off as a tree search problem.}
    \label{fig:beam-search}
    \Description{Contrast between repair-only, replace-only, and beam search for refining program candidates.
The figure represents different strategies for refining program candidates through a tree search paradigm. The “repair-only” approach (depth-first) follows a linear path, iteratively debugging a single candidate. The “compromise” approach (beam search) maintains a limited pool of candidates, balancing repair and replacement to explore multiple potential solutions. The “replace-only” approach (breadth-first) generates entirely new candidates at each step, discarding previous ones. The diagram visualizes these strategies as search trees, with beam width controlling the number of candidates retained. Repair-focused methods prioritize incremental improvement, while replacement-oriented methods favor broader exploration.}
\end{figure}

After evaluating a given candidate solution in \execute{}, \method{} supports two approaches to address the candidate's flaws:
\begin{itemize}
\setlength{\parskip}{0pt}
\setlength\itemsep{0pt}
  \item \emph{Replace} the candidate with another sample from the current population.
  \item Use \instruct{} and \debug{} to repair the candidate.
\end{itemize}
We refer to this problem as the \emph{repair-replace trade-off}, by analogy with production economics~\cite{jack2000:optimal}. 

How does the choice of hyperparameters $\treearity{},$ the total number of candidate programs in each generation, and $\beamwidth{},$ the number of selected repairs to be preserved in a generation, influence the flow of \method{}?
$\treearity$ and $\beamwidth{}$ act as upper bounds on the \emph{replace} option by limiting the size of the population.
In the edge cases, $\treearity{} = \beamwidth{} = 1$ corresponds to a repair-only process, while $\treearity{} = \beamwidth{} = \infty$ corresponds to replace-only, as illustrated in Figure~\ref{fig:beam-search}. 
Here, the repair-only scenario can also be seen as an LLM-guided random walk~\cite{xia2020:random} and replace-only as random sampling from the LLM.
Strategies with tree arities between $1$ and $\infty$ are similar to population-based evolutionary algorithms.
Note that $\treearity{}$ is defined by $\treearitydraft{}$ for the initial draft solutions in the first generation and $\treearityexplain{} \cdot \treearitydebug{}$ for later generations. 

Observe that a mutation-only genetic algorithm with tournament selection with fixed population size $\beamwidth{},$ such as \method{}, is equivalent to \emph{local beam search} with beam width $\beamwidth{}$ on an $\treearity{}$-ary tree ~\cite[Section 4.1.4]{russell2010:artificial}. This corresponds to a known property of local beam search: it degenerates into a depth-first search at $\beamwidth{} = 1$, whereas setting $\beamwidth{} = \infty$ yields a breadth-first search.

\section{Related Work}
\label{sec:related-work}

Until recently, the tasks of neural program synthesis~\cite{gulwani2017:program} and program repair~\cite{legoues2019:automated, petke2018:genetic, legoues2012:systematic,zhang2024systematic} have been considered separately.
However, results from genetic programming~\cite{sobaniaRecentDevelopmentsProgram2021} suggest that evolution is a crucial step in synthesis.
A number of important studies bridging this gap in the application of large language models have been carried out concurrently with this work, discussed below.

The use of large language models for a program repair step within a program synthesis pipeline has been studied by \citet{joshi2022:repair:arxiv} and \citet{gupta2020:synthesize}, 
while the specific case of instruction-driven LLMs has been explored by \citet{fan2023:automated}, where the initial synthesis is done by the Codex model~\citep{chen2021:evaluating} and for program repair both Codex and state-of-the-art specialized repair tools such as TBar and Recoder~\citep{just2014:defects4j} are considered and compared. 
\citet{zhang2023:selfedit} do the same, but fine-tune PyCodeGPT-110M~\citep{zan2022:cert} to use it as a repair model. 
The resulting framework is a two-step process (1 draft step and 1 debug step), while iterative evolution and search are not explored. 

Evolution through Large Models (ELM)~\citep{lehman2022:evolution} proposes to use a language model in place of a mutation operator within a traditional genetic programming framework~\citep{koza1994:genetic}. They use a type of instruction fine-tuned model trained on git commit messages known as a diff model.\footnote{~\url{https://carper.ai/diff-models-a-new-way-to-edit-code/}}
However, the model is directed neither to solve the programming problem at hand nor to fix bugs.
Instead, it is provided with generic instructions such as ``Change function f.'' 
This approach is meant for the cases where creative freedom~\citep{stanley2015:why} is encouraged rather than satisfying concrete requirements.
\citet{liu2023algorithm} demonstrate the advantages of applying ELM to modular components of a solution, rather than the entire solution to the given problem.

Some related work is explicitly guided by the metaphor of conversation between agents. 
\citet{zhang2023steam} implements a repair-only program synthesis loop as a conversation between an analyst agent, a coder, and a tester, while \citep{dong2024:selfcollaboration} does the same for tester, developer, and reviewer. 
The coder and the developer are roughly equivalent to \synthesize{}, the tester to \execute{}, and the reviewer to \instruct{}, while the analyst is an agent that prepares a generation prompt for the coder. 
Both testers are language models that predict the output of a program without actual compilation, execution, or testing, which makes \citet{dong2024:selfcollaboration} and \citet{zhang2023steam} specific cases of chain-of-thought prompting~\cite{yu2023:better} for program synthesis.

Several studies explore an iterative approach to program synthesis in a manner similar to \method{}~\citep{xia2023:conversational,chen2023:teaching,shinn2023:reflexion}. 
However, they do not explore the repair-replace trade-off and exclusively implement the repair-only approach that is prone to local minima.
SelfEvolve~\citep{jiang2023:selfevolve} is a repair-only version of a framework similar to \method{}.
SeflEvolve demonstrates the benefits of LLMs evolving not just the source code but an additional natural language text file that acts as the system's knowledge base and is included in the model prompt when generating code. 
Finally, Self-Taught Optimizer (STOP)~\cite{zelikman2023:selftaught} takes the concept of self-improvement to the meta level and uses a large language model to edit the evolutionary algorithm itself (i.e., blocks or contents of Figure~\ref{fig:method}). 
Their reflections on the safety implications of such automated algorithm changes are of particular interest when considering this trajectory~\cite[Section 8]{zelikman2023:selftaught}. These concerns do not hold in the context of \method{} because the algorithm is fixed.
In other words, \method{} does not self-evolve but creates solutions that it improves, and the solutions that are synthesized are closely constrained by the validation set used by the \execute{} agent.

These efforts to combine large language models and evolutionary techniques fall within the broader context of the quest for effective inductive bias in genetic programming~\citep{whighamSearchBiasLanguage1996}: a crucial property for any learning algorithm~\citep{haussler1988:quantifying}. 
\citet{reuterGraphNetworksInductive2023} suggest using graph neural networks for this purpose, while grammatical evolution methods, such as WHGE~\citep{bartoliWeightedHierarchicalGrammatical2020} and $\pi$-GE~\citep{oneill2004:pgrammatical} use the known grammar of the programming language.
Large language models are a novel and promising~\citep{custodeComparingLargeLanguage2024} alternative to grammatical evolution that incorporates semantics and idiom~\citep{allamanisMiningIdiomsSource2014,orlovFindingIdiomsSource2020} of a programming language in addition to its grammar. 

\section{Experimental Design}
\label{sec:eval}

To explore the capabilities of \method{} and its generalizability, we test the framework on two benchmarks (PSB2 and
HumanEval-X), a total of two different ranking strategies (tournament selection and lexicase selection),
three models in the coding part of SEIDR (Codex, GPT-3.5, and Llama 3), two programming languages (Python and C++), and various branching factors.
We use three models over the two parts of our experiments: one model in the initial exploration (also reported by~\cite{liventsev2023:fully}) and two models in the generalizability part.
The problems in the benchmarks originate from coding competitions and human-written programming assignments. 

During our empirical evaluation of \method{}, we address the following research questions:
\head{\rqtreearity{}. Repair-replace trade-off exploration} 
What is the impact of using different tree search strategies
in the autonomous programming setting? 
We experiment with six different tree arities but fix the tournament selection in the ranking part and one prompt. 
Here, we study the impact of tree arity on the number of resolved problems as well as the speed of obtaining solutions.   
\head{\rqllama{}. Generalizability of the approach to different LLMs and an additional dataset} 
How does the choice of an LLM affect the performance of SEIDR? 
We vary the tree arity and experiment with two additional LLMs and one additional dataset.
By default, we use tournament selection as the ranking strategy. 
\head{\rqmultirun{}. Repeatability of \method{} in multiple runs with the same hyperparameters} 
How does the non-deterministic nature of LLMs affect \method{} performance when the method is restarted several times with the same hyperparameters?
We study how \method{} results vary in different restarts of the same experiments with unchanged hyperparameters as a result of the LLM non-determinism. 
The motivation for \rqmultirun{} is that LLMs exhibit stochastic behavior: the same prompt can yield different responses. 
Essentially, LLMs generate answers token-by-token and predict the next tokens based on the probability distribution over a vocabulary of tokens, which is sensitive to the precision of floating-point operations. 
If two tokens are predicted to be the next ones, with a very similar probability, it is likely that either of them will be chosen at each individual run. 
Further tokens are generated auto-regressively and depend on the previous tokens, so once one token diverges, the whole sequence is likely to diverge, too. 
\head{\rqlexicase{}. Effect of changing the parent selection strategy in the \rank{} agent}
How does the lexicase selection-based ranking strategy impact performance in comparison to tournament selection in the \rank{} agent? 
We use the best-performing tree arities from \rqmultirun{} to run the experiments with lexicase selection as the ranking strategy instead of tournament selection, to explore whether a different parent selection algorithm can further improve the results.

\subsection{Data}
\label{sec:data}

Our experiments use the Program Synthesis Benchmark~2 (PSB2)~\cite{helmuth2022:applying} and HumanEval-X~\cite{zheng2023:codegeex} in C++ and Python. 
The key criteria for this choice are the availability of task descriptions in English and unit tests in Python and C++ or language-agnostic unit tests as well as the wide acceptance of these benchmarks in the areas of generative LLMs (HumanEval) and genetic programming (PSB2). 

\subsubsection{PSB2}
The first dataset is a benchmark suite of 25 problems for program synthesis that resemble small real-world tasks. PSB2 was developed as a more realistic and challenging version of PSB1~\cite{helmuth2015:general}, the latter consisting of textbook problems and is widely used in genetic programming~\cite{sobania2022:choose}. 
The problems require different data structures and control flows to be used for effective solutions and are taken from sources, such as competitive programming platforms and educational courses. 
The problems have descriptions in English, as well as 1 million~(M) tests for training and 1M testing-stage tests, including edge or corner cases that test the resulting program on complicated inputs. 
The tests are provided as I/O pairs and are distributed together with the problem descriptions as a PyPI package.\footnote{~\url{https://pypi.org/project/psb2/}} 

In PSB1, the training set consists of the edge test cases and is augmented by random test cases if the number of edge tests is not enough. The test set is formed by random test cases. 
This terminology is preserved in PSB2.
We use the PSB2 training set for ranking and selection of programs (validation) within an experiment and the test set for reporting the result thereof (testing).
Thus, we will refer to the PSB2 training set as the \emph{validation set}, to be more consistent with how it is used in \method{}.

\subsubsection{HumanEval-X}
The second dataset that we use is a development from the original set of human-written programming tasks in HumanEval~\cite{chen2021:evaluating}, which is a standard code generation benchmark for LLMs.
HumanEval consists of 164 problems with a docstring representing problem description, a function signature, a correct solution, and unit tests in Python. 
HumanEval-X is the result of translating correct HumanEval programs and unit tests into five programming languages~\cite{zheng2023:codegeex}. 
We use HumanEval-Python for experiments in Python to ensure a comparison with other models in the setup without \method{}. 
In addition, we test \method{} on the HumanEval-C++ part of HumanEval-X. %

The test functions of HumanEval-X contain all tests in one function. We split the aggregated test functions into separate tests so that the \rank{} agent can evaluate the \text{score}. 
On average, the number of tests in HumanEval-Python is 7.25 and 6.95 in HumanEval-C++, which is appointed to a repeated additional test present in some HumanEval-Python examples of the following type: \texttt{assert True, "This prints if this assert fails 1 (good for debugging!)"}.
This test type is not present in HumanEval-C++.

To reiterate, we keep the original HumanEval-Python setup for direct comparison with the models tested on this benchmark without \method{}. 
Because of the limited number of tests, we pass up to five tests to the draft prompt and make all tests visible to \method{} for the debugging loop. 
In other words, we do not have a held-out test split for HumanEval-X in the same manner as we do for PSB2.

\subsection{Models}
\label{sec:models}

\method{} uses up to three LLMs --- \synthmodelnoargs{}, \textmodelnoargs{}, and \debugmodelnoargs{} ---
in \synthesize{}, \instructllm{}, and \debug{}, respectively. 
The main prerequisite is that \synthmodelnoargs{} and \debugmodelnoargs{} are a text-to-code models which take both a textual description and a draft code as input.
Therefore, \synthmodelnoargs{} and \debugmodelnoargs{} can be a chat model, a code completion model, an instruction fine-tuned, or a foundation generative language model pre-trained on code in addition to text. 
By analogy, the text-to-text \textmodelnoargs{} can be a chat, instruction fine-tuned, a text completion, or a text generation model pre-trained on text and code.

In our experiments, we use Open AI Generative Pre-trained Transformer (GPT) models by Open AI and Llama 3 by Meta~\cite{roziere2023:code}. 
GPT models are auto-regressive transformer models that have the decoder-only architecture as opposed to the original full encoder-decoder transformer.
They are pre-trained on both text and code and excel at sequence-to-sequence generative tasks, including code-to-code, text-to-code, and code-to-text.

In our initial experiments, we use Codex-edit\footnote{~\href{https://openai.com/index/gpt-3-edit-insert/}{code-davinci-edit-001}} 
as the LLM for writing and debugging programs and GPT-3\footnote{~\href{https://platform.openai.com/docs/deprecations}{text-davinci-003}} for bug summarization via text completion~\cite{brown2020:language} --- both being 175B-parameter models.
In our generalizability experiments, we use the GPT-3.5 model\footnote{~\href{https://platform.openai.com/docs/models/gpt-3-5-turbo}{gpt-3.5-turbo}} for program synthesis, bug summarization, and debugging. 
The GPT-3.5 model is an improvement over GPT-3 that is optimized for chat and available through an API.
This switch is mainly motivated by rapid model updates, an OpenAI announcement that GPT-3 was due to become obsolete, i.e., not actively supported by the company, along with the company's recommendation to switch to GPT-3.5. 

To further evaluate the generalization capabilities of \method{}, we have chosen Llama 3-8B~\cite{roziere2023:code}, an open-source alternative to GPT models with the same standard decoder-only transformer architecture. 
Compared to Llama 2~\cite{touvron2023:llama}, Llama 3 introduces improvements to the architecture and the training process, such as grouped query attention.

\subsection{Prompts}
\label{sec:prompts}

\subsubsection{Prompts in the Initial Exploration of SEIDR}
\label{sec:prompt-strategies}

The prompt for the LLM model is static and consists of the input for editing --- candidate program generated so far --- and a debug instruction to repair the candidate. 
The debug instructions are formulated as templates. The instructions describe the violated requirements in terms of the wrong output in a failing I/O test or summarize the bug to capture issues in code logic.
We present debug instructions using the template engine format: the brackets \{ \} denote that the placeholder in the brackets will be replaced with the value generated during execution, \{I$_{\text{val}}$\} and \{O$_{\text{val}}$\} stand for values of the validation set I/O pair. As shown in Figure~\ref{fig:method-instruct}, the instruction to fix execution errors that abort the program before the resulting output is obtained with \texttt{stderr} lines: Fix \{stderr\}. Debug instruction that uses the output of candidate program execution is static, formulated as follows: 
\begin{equation}\label{seidr:prompt-0} 
    \text{\smalltt{Make sure that I}}_{\text{val}} \text{\smalltt{ -> O}}_{\text{val}}. \tag{S0}
\end{equation}

\subsubsection{Prompts for Instruction Fine-tuned and Chat Models}
\label{sec:ollama-prompts}

The prompts presented in this section are used in the experiments with GPT-3.5 and Llama 3.
This implementation is optimized for instruction and chat models, which use prompts as inputs represented as text, partially with code fragments.
The models use a system message that describes the ``role'' of the LLM and a regular message that works as an instruction or a chat message from a user.
To provide more context, we always use a problem description, problem name, and programming language to the model as textual input (\texttt{descr}). 
As code, we also add an initial template depicted in Figure~\ref{fig:template} to \synthmodelnoargs{} and the current program candidate to \textmodelnoargs{} and \debugmodelnoargs{}. The resulting prompts are as follows:
\newline\newline
The system message is: \newline
\begin{tabular}{p{.975\linewidth}}
\midrule
\smalltt{You are an experienced software developer.} \\
\smalltt{You write concise code in \{language\}.} \\
\smalltt{The code must read input from user and return output corresponding to the task description.} \\
\midrule
\end{tabular}
\newline\newline
The input to \synthmodelnoargs{} looks as follows:\newline 
\begin{tabular}{p{.975\linewidth}}
\midrule
\smalltt{Solve the following code contest problem: \{problem\_name\}.} \\
\smalltt{Problem description: \{problem\_description\}.} \\
\smalltt{\{program\_template\}} \\
\smalltt{Only complete the code, do not add triple quotes, do not give explanations.} \\
\midrule
\end{tabular}
\newline\newline
Bug explanations are generated with \textmodelnoargs{} using the following instructions:\newline
\begin{tabular}{p{.975\linewidth}}
\midrule
\smalltt{I'm trying to solve the following code contest problem: \{problem\_name\}.} \\
\smalltt{Problem description: \{problem\_description\}.}\\
\smalltt{Currently, the code is} \\
\smalltt{\`{}\`{}\`{}} \\
\smalltt{\{program\_candidate\} }\\
\smalltt{\`{}\`{}\`{}} \\
\smalltt{The issue is} \\
\smalltt{\{stderr\} or "it must return \{expected\_output\} for input \{input\},} \\
\smalltt{but it returns \{output\}".}\\
\smalltt{Describe how I should fix the code in a very concise manner.} \\
\midrule
\end{tabular}
\newline\newline
And the debugging model \debugmodelnoargs{} operates on the following instruction:\newline
\begin{tabular}{p{.975\linewidth}}
\midrule
\smalltt{Solve the following code contest problem: \{problem\_name\}.} \\
\smalltt{Problem description: \{problem\_description\}.} \\
\smalltt{Currently, the code is } \\
\smalltt{\`{}\`{}\`{}} \\
\smalltt{\{program\_candidate\}} \\
\smalltt{\`{}\`{}\`{}} \\
\smalltt{Modify the code as \{bug\_summary\}.} \\
\smalltt{You must only return correct code. } \\
\smalltt{Remove any triple quotes, language name or explanations.}\\
\midrule
\end{tabular}

\subsection{Repair-replace Trade-off Settings}
\label{sec:trade-off-settings}

The settings for tree arity will also be divided into two experiment sets: the ones for GPT-3 and Codex, and the ones for the GPT-3.5 and Llama 3 experiments.

\subsubsection{Tree Arity for the Initial Exploration of SEIDR}
\label{sec:tree arity-gpt-3}
As described in Section~\ref{sec:beam-search}, the population size (number of parents to choose from in the tournament selection or beam width from the beam search perspective) $\beamwidth{}$ and tree arity $\treearity{}$ define the repair-replace trade-off, where higher $\beamwidth{}$ and $\treearity{}$ correspond to repair over replace. 
We evaluate four options for these hyperparameters as shown in Table~\ref{tab:seidr:w-n-initial-exploration}. 
We only run the experiments once, due to the experimental timeline and the discontinuation of model support by the model provider. 

\begin{table}[t]
\setlength{\tabcolsep}{20pt}
\centering
\caption{Initial exploration of \method{}: hyperparameters in the tree arity experiments.}\small
\label{tab:seidr:w-n-initial-exploration}%
\begin{tabular}{rcccc}
\toprule
experiment \# & 1 & 2 & 3 & 4 \\
\midrule
population size (beam width), $\beamwidth{}$ & 1 & 10 & 100 & $\infty$ (1000) \\[1pt]
tree arity, $\treearity{}$ & 1 & 10 & 100 & $\infty$ (1000) \\[1pt]
\midrule
max programs generated & \multicolumn{4}{c}{1000} \\[1pt]
prompt & \multicolumn{4}{c}{\ref{seidr:prompt-0}} \\[1pt]
models  & \multicolumn{4}{c}{\parbox{5cm}{\centering Codex as $p_\text{synth} \text{ and } p_\text{debug}$ 
}} \\[1pt]
\midrule
\parbox{4cm}{\raggedleft \# restarts (or runs) \\ with the same hyperparameters} &  
\multicolumn{4}{c}{1} \\[8pt]
datasets  & \multicolumn{4}{c}{PSB2} \\[1pt]
languages  & \multicolumn{4}{c}{Python, C++} \\
\bottomrule
\end{tabular}
\end{table}

Because we aim to compare tree search parameters, we fix one default debugging instruction~\ref{seidr:prompt-0} and use the \instructs{} agent.  
Moreover, we set the upper limit for the total number of generated program candidates to 1000 to limit the experimentation time. 
Although some solutions may not be found within the hard limit, we assume\footnote{~This assumption is later confirmed in Section~\ref{sec:seidr:rqtreearity}.} that 1000 program candidates form a sufficiently large search space for our experiments.
$\beamwidth{} = \treearity{} = \infty$ is achieved in implementation by setting equal $\beamwidth{}$ and $\beamwidth{}$ equal to the upper limit of the program count of 1000.
This ensures that a second generation of programs does not exist.

\subsubsection{Tree Arity for SEIDR Generalizability Experiments}
\label{sec:tree arity-ollama} 
With the shift to chat and instruction models in the generalizability part of our study, we move from generating one bug explanation and one code draft or update to a batch of those. 
Specifically, each of the three LLMs in \synthesize{}, \instruct{}, and \debug{}  can generate sequences in batches. 
We generate $\treearitydraft{}$ programs in the first generation with \synthmodelnoargs{} model, $\treearityexplain{}$ bug explanations with \textmodelnoargs{} for each program in a generation, and $\treearitydebug{}$ candidate repairs for each of the debugging instructions using \debugmodelnoargs{}.
A new generation of $\treearityexplain{} \cdot \treearitydebug{} \cdot \beamwidth{}$ programs created from each of $ \beamwidth{}$ parents in a previous generation is ranked and filtered to keep the best-performing $\beamwidth{}$ candidates for generating the next candidates. 

To balance between a reasonable number of experiments and diverse sets of hyperparameters, we fix $\treearityexplain{}=2$ to moderately vary the bug descriptions and set $\treearitydraft{} = \treearitydebug{} = \treearity{}.$
As a reference, in the experiments with GPT-3 and Codex, we generated only one bug explanation ($\treearityexplain{} = 1$) and used $\treearitydraft{} = \treearitydebug{} = \treearity{}$ setting, too. 
We evaluate six options of $\treearity{}$ 
as shown in Table~\ref{tab:w-n-generalizability} and use tournament selection as the ranking strategy. 

The choice of these tree branching hyperparameters and the maximum number of generated programs is motivated by the experiments with GPT-3 and Codex, where the best results were obtained for $\treearity{}=10.$ 
Therefore, we explore the area around this value more closely in the generalizability experiments.
In the same experiments, the majority of problems in PSB2 were solved within the first 100 generated programs.
Therefore, the upper limit for the total number of generated program candidates is set here to 100 to limit the experimentation time.

To account for the stochasticity of language models and the fact that OpenAI's models do not support setting the effective sampling temperature to zero to force deterministic behavior,\footnote{~\url{https://community.openai.com/t/observing-discrepancy-in-completions-with-temperature-0/73380}} we ran the experiments six times with each set of hyperparameters.
This number of runs was selected to hit a sweet spot between the overall running time and costs of the experiments, while at the same time achieving confidence in the stability of the results in the presence of non-determinism. 
Note that reporting results over six runs is considerably better than the common practice of having only one run for every selection of hyperparameters, and it is in line with the best-of-class practice in the field of LLMs for code generation~\cite{ouyang2023:llm}.

The total cost of running the experiments with GPT-3 and Codex were around 550 USD, and the experiments with \gpt{} amounted to 266 USD.\footnote{~For comparison, one run with GPT-4o cost us 315 USD (early July 2024), so further use of this model was discarded.}
Moreover, the time to finish one run with several branching factors and all the tests amounts to ca. 42h for PSB2\footnote{~Due to the local setup for testing, API call limits, and the number of tests.} and ca. 156h for HumanEval-X.
Overall, we restart experiments with GPT-3.5 six times for each of the tree arities $ N_{\text{synth}} = N_{\text{debug}} = N^* \in \{1,2, 4,10,16,100\}$ and the same for Llama 3, with a total of $6 \times 6 \times 2 = 72$ experiments. 
For the lexicase selection experiments, we have 6 runs per model but one best-performing tree arity, which adds $12$ experiments to the total count.

\begin{table}[t]
\setlength{\tabcolsep}{10pt}
\centering
\caption{\method{} generalizability experiments: hyperparameters in the tree arity grid search.}\small
\label{tab:w-n-generalizability}
\begin{tabular}{rcccccc}
\toprule
experiment \# & 16 & 17 & 18 & 19 & 20 & 21\\
\midrule
population size (beam width), $\beamwidth{}$ & 1 & 4 & 8 & 10 & 16 & $\infty$ (100) \\[4pt]
\# programs in the 1st generation, $\treearitydraft{}$ & 1 & 4 & 8 & 10 & 16 & $\infty$ (100) \\[4pt]
\# bug explanations for candidate, $\treearityexplain{}$ & 2 & 2 & 2 & 2 & 2 & - \\[4pt]
\# repairs for each explanation, $\treearitydebug{}$ & 1 & 4 & 8 & 10 & 16 & - \\[4pt]
\midrule
max programs generated & \multicolumn{6}{c}{100} \\[4pt]
prompts & \multicolumn{6}{c}{see Section~\ref{sec:ollama-prompts}} \\[4pt]
models  & \multicolumn{6}{c}{
 \parbox{5cm}{
     (a) GPT-3.5 as $p_\text{synth,} \; p_\text{debug,} \; p_\text{explain,}$ \\
     (b) Llama 3 as $p_\text{synth,} \; p_\text{debug,} \; p_\text{explain}$
     }
} \\[10pt]
\midrule
\# runs per experiment &  \multicolumn{6}{c}{6} \\[4pt]
datasets  & \multicolumn{6}{c}{PSB2, HumanEval-X} \\[4pt] 
languages  & \multicolumn{6}{c}{Python, C++} \\[4pt]
\bottomrule
\end{tabular}
\end{table}

\subsection{Performance Indicators}
\label{sec:metrics}

\sloppy %
In our experiments, we compare 
the number of fully solved programs obtained with \method{} with different values of hyperparameters. 
For a more detailed analysis of results, we use \emph{test pass rate (TPR)} and \emph{Excess Programs Generated (EPG)}.
TPR reflects the percentage of fully passed test cases based on the exact match of program output and test output. 
The TPR metric is used for the final evaluation of generated programs and does not reflect partial passing of the I/O test as opposed to the \emph{score} as calculated by the \rank{} agent (see Section~\ref{sec:execute}). 

We define \emph{pass@k} as the number of problems that have $TPR=1$ if SEIDR is stopped after generating $k$ programs, following~\citet{kulal2019:spoc} \emph{``success rate at budget of $k$ programs.''}
Note that \citet{jiang2023:selfevolve} and \citet{chen2023:teaching} define $k$ as the number of restarts of the iterative method, the budget in terms of trees of programs.
We choose against this approach, since it threatens the validity of the comparison between iterative tree-based program synthesis and repair-only baseline by giving the iterative approach additional budget in terms of the number of programs it can generate.

Codex \cite{chen2021:evaluating} calculates pass@n>k and constructs an unbiased estimator of pass@k with lower variance, ensuring statistically robust results.
We cannot apply this adjustment for \method{}, since the adjustment assumes that the programs are independent and identically distributed, while \method{} is a Markov chain with dependencies between iterations. 

EPG reflects the number of programs generated before the first occurrence of the program that passes all validation test cases.
EPG is indicative of the computational cost of solving a problem distributed in terms of LLM inferences and program compilations and executions.
For a single execution of SEIDR, EPG is equivalent to the smallest $k$ at which pass@k=1.

\subsection{Implementation Details}
\label{sec:implementation}

To summarize the setup, in this study, we have two groups of experiments. 
The first group of experiments is dedicated to the initial exploration of \rqtreearity{}
with Codex-edit (code-davinci-edit-001) as the LLM for writing and debugging programs. 
We have referred to these experiments as \emph{Initial Exploration of \method{}} with Codex and GPT-3, and test the hyperparameter choices only on PSB2 as detailed in Table~\ref{tab:seidr:w-n-initial-exploration}.
Here, we use wide steps between tree arity values (see Section~\ref{sec:tree arity-gpt-3}).
 
The second set of experiments mainly focuses on the generalizability (\rqllama{}) of \method{}
and its robustness to restarting experiments with the same hyperparameters (\rqmultirun{}).
The motivation here is to potentially improve on GPT-3 with a newer, generally more powerful version, GPT-3.5, and its smaller open-source competitor, Llama 3.
GPT-3.5 and Llama 3 are used in more fine-grained repair-replace trade-off exploration and ranking experiments (\rqtreearity{}). 
We have referred to these experiments as \emph{\method{} Generalizability Experiments} and test \method{} both on PSB2 and HumanEval.
Building on the findings of the initial exploration, we use more fine-grained tree arity values (see Section~\ref{sec:tree arity-ollama}, Table~\ref{tab:w-n-generalizability}) and use the prompts from Section~\ref{sec:ollama-prompts}. 
Thus, \instruct{} is represented by the \instructllm{} agent and creates $\treearitydebug{}$ bug summaries.
Each program update creates $\treearity{}$ child programs from one parent with the \synthesize{} and \debug{} agents.
We also compare the performance of \method{} with the current state-of-the-art without \method{} (see Section~\ref{sec:results-rqllama}).

The second set of experiments is further updated with an alternative ranking strategy, lexicase selection (\rqlexicase{}). 
For each model, dataset, and language, we choose the best-performing tree arity from \rqllama{} and exchange the tournament selection algorithm with the lexicase selection. 
This selection step chooses parents for debugging updates in each generation. 

In all experiments, we set the limit to generate a maximum of $M$ program candidates during the search for the candidate that passes all validation tests. 
If we reach $M$ candidates and none of them pass all validation tests, we store the test pass rate for the last generated candidate and the best test pass rate achieved throughout the search. 
For the first set of experiments, we set $M = 1000,$ and for the generalizability ones, we limit $M$  to $100,$ after finding out that for the majority of problems, a solution is found among the first 100 programs or not found at all.

Following~\citet{helmuth2021:psb2}, we use 2000 I/O pairs ($\left(I, O\right)_{test}$ in Figure~\ref{fig:method}) from the test split of PSB2 to evaluate the candidate program that has passed all validation test cases ($\left(I, O\right)_{val}$ in Figure~\ref{fig:method}) during debugging. 
Due to repetitive calls to \execute{}, we have to resolve the speed of testing versus precision trade-off while choosing the number of validation test pairs.
We resolve the trade-off by fixing the validation set size at 100 for the initial experiments and 50 for the generalizability exploration, which has more runs with the same hyperparameters. 
We have run a preliminary experiment to confirm that we do not lose the final test pass rate points on 2000 tests when we decreased the validation test set size from 100 (which was used in the GECCO-2023 paper) to 50 for the generalizability exploration.
Due to a small number of tests in HumanEval-X, all tests are made visible to the debugging LLM and during the validation step.  
To operate with the chosen LLMs in \method{}, we use ollama\footnote{~\url{https://ollama.ai/}} and LangChain.\footnote{~\url{https://www.langchain.com/}}  
To ensure that the program candidates generated from the same parent program are different from each other, we change the temperature parameter of the LLMs.

\section{Results and Discussion}
\label{sec:results}

In this section, we present the results of the initial exploration, where we investigate the repair-replace trade-off in \method{} with Codex and GPT-3 (\rqtreearity{}) using the PSB2 benchmark.
We then continue with generalizability (\rqllama{}) and repeatability (\rqmultirun{}) experiments with GPT-3.5 and Llama 3 on PSB2 and HumanEval.
Finally, we test lexicase selection as the ranking strategy (\rqlexicase{}).

\subsection{Initial Exploration}

\subsubsection{Repair-replace Trade-off in the Initial Exploration of SEIDR}
\label{sec:seidr:rqtreearity}
\begin{figure}[t]
  \centering
  \includegraphics[width=0.5\linewidth, trim={0mm 2.8mm 0mm 2mm}, clip]{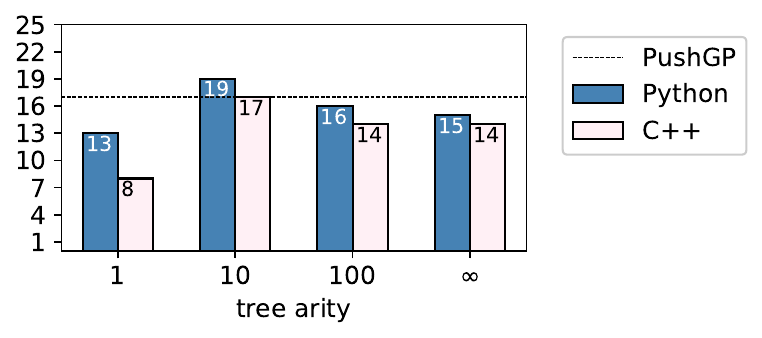}
  \caption{Number of solved PSB2 problems depending on tree arity in beam search for prompt type \ref{seidr:prompt-0}.}
  \label{fig:seidr:solved-vs-bf}
  \Description{Solved PSB2 problems based on tree arity in beam search.
The figure shows a bar chart comparing the number of PSB2 problems solved using beam search with different tree arity values (1, 10, 100, and ∞) for Python and C++. Python results are shown in blue, while C++ results are in pink. A dotted horizontal line represents the performance of PushGP. The results indicate that increasing tree arity generally improves problem-solving performance, with the best results observed at tree arity 10. Performance declines at higher tree arity, suggesting diminishing returns for broader exploration.}
\end{figure}

We compare the number of solved problems in the experiments with tree arity of 1, 10, 100, and $\infty$ and fixed debug instruction \ref{seidr:prompt-0} in Python and C++ in Figure~\ref{fig:seidr:solved-vs-bf}. 
The results of \method{} are compared to the baseline performance of PushGP on the PSB2 benchmark, which solves 17 out of 25 problems. 
Note that experiments with $N=1$ and $N=\infty$ can be considered as ablation studies, where the replace option and repair option are turned off correspondingly. 

The results highlight the benefit of compromise strategies with tree arity of 10 and 100 over repair-only ($N=1$) and replace-only ($N=\infty$) strategies. 
The results show that the repair-only scheme is outperformed by other strategies. 
We explain the poor performance of the repair-only strategy by the fact that the search space is under-explored. 
Specifically, the replace scenario ensures that the LLM for the debugging component represented by Codex-edit in our experiments generates different updates of program candidates using variable temperatures.
The probability of finding a better fix is higher when more alternatives are generated to update the draft program at $N>1$ compared to $N=1$. 
The search strategy with $N=10$ yields the best results: it performs on par with PushGP for C++ and outperforms the baseline during Python program synthesis by +2 problems, resulting in a total of 19 programs that pass all test cases.
The results imply that generating a moderate number of programs in parallel during the \debug{} step works better than the policies in which more updates are generated for each program (100 or 1000), 
or those in which only one program is updated iteratively.

\begin{figure}[t]
\begin{subfigure}[t]{\columnwidth}
\centering
\includegraphics[width=.7\linewidth, trim={0mm 4mm 0mm 0mm}]{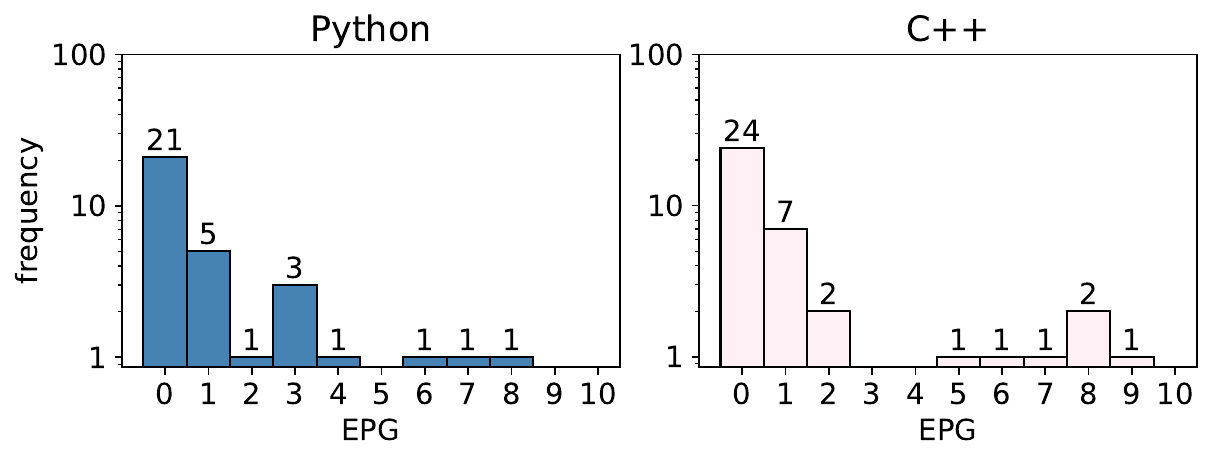}
  \caption{0 $\leq$ EPG $\leq$ 10 with step 1.}
  \label{fig:seidr:epg-distrib-solved-10}
\end{subfigure}

\begin{subfigure}[t]{\columnwidth}
\centering
\includegraphics[width=.7\linewidth, trim={0mm 4mm 0mm 0mm}]{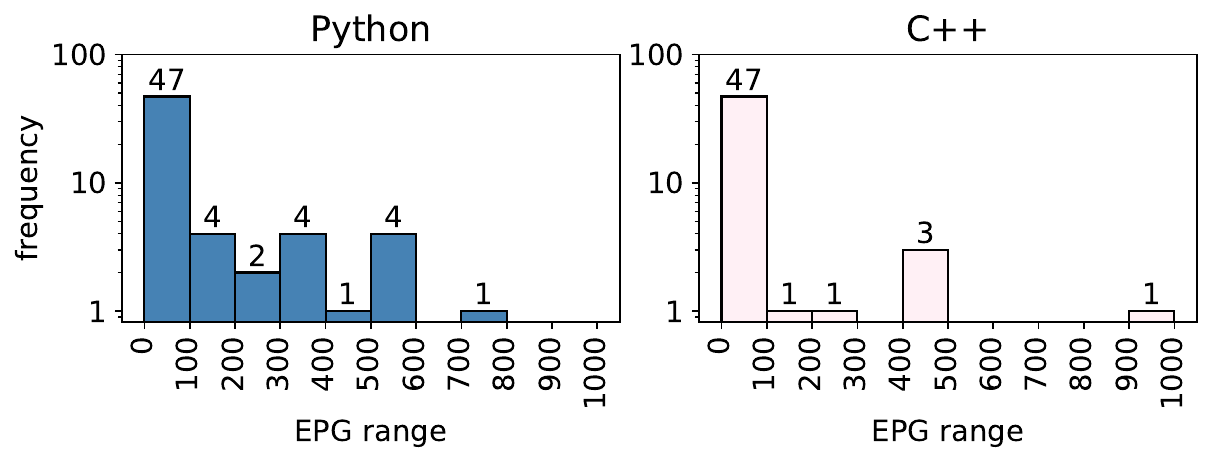}
  \caption{0 $\leq$ EPG $\leq$ 1000 with step 100.}
  \label{fig:seidr:epg-distrib-solved-100}
\end{subfigure}
\caption{Distribution of the number of generated programs during each problem-solving attempt in the experiments with different tree arities where a problem solution is found.}
\label{fig:seidr:epg-distribution}
\Description{Distribution of generated programs per problem-solving attempt.
The figure consists of four histograms comparing the number of generated programs during problem-solving attempts for Python and C++. The top row presents the distribution for execution per generation (EPG) values between 0 and 10 with a step size of 1, while the bottom row extends the range from 0 to 1000 with a step size of 100. The y-axis represents the frequency of occurrences on a logarithmic scale. Both Python and C++ exhibit a high concentration of attempts with low EPG values, particularly at EPG = 0. However, Python shows a more gradual decrease in frequency as EPG increases, whereas C++ exhibits a sharper drop-off. The results suggest that successful problem-solving often requires fewer program generations, with a small subset of cases requiring significantly higher EPG values.}
\end{figure}

\begin{figure*}[t]
  \centering
  \includegraphics[width=.56\textwidth, trim={3mm 1.6mm 3mm 2mm}, clip]{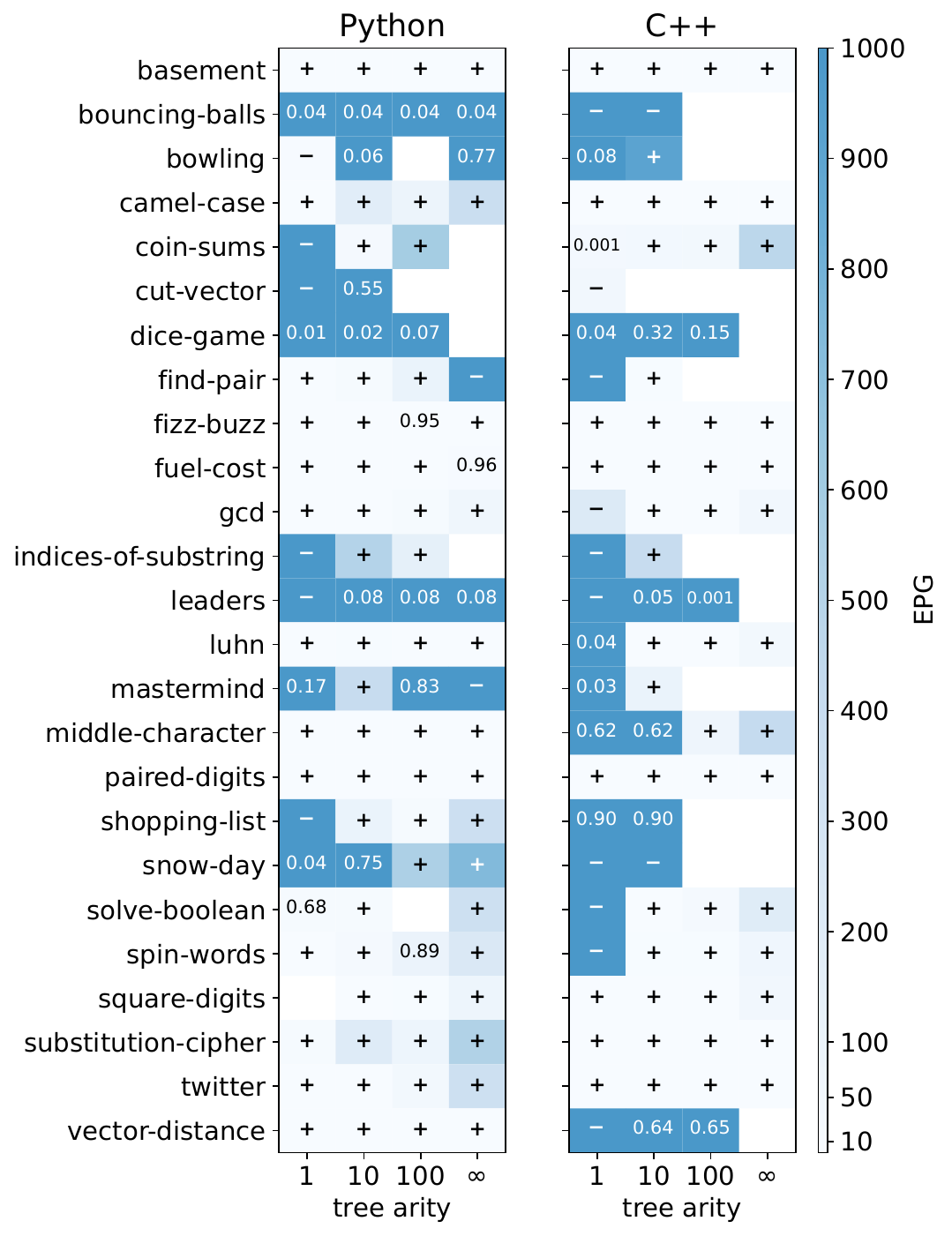}
  \caption{Number of excess programs generated (in color) and test pass rate (as numbers) depending on tree arity. Higher EPG values are shown in darker shades. We denote solved problems with ``+'' (test pass rate = 1), unsolved problems with ``-'' (test pass rate = 0), and show the test pass rate for partially solved problems. }
  \label{fig:seidr:epg-bf}
  \Description{Excess programs generated and test pass rates across tree arities.
The figure presents a heatmap comparing the number of excess programs generated (EPG) and test pass rates for Python and C++ across different tree arities (1, 10, 100, and ∞). Each row corresponds to a benchmark problem, with solved problems marked by “+” (pass rate = 1), unsolved problems by “−” (pass rate = 0), and partially solved problems showing their pass rates as numeric values. Darker shades indicate higher EPG values. The results suggest that increasing tree arity affects problem-solving efficiency, with some problems showing improved test pass rates as tree arity grows, while others remain unsolved regardless of the search strategy. Python and C++ exhibit different patterns in EPG distribution, indicating language-specific variations in program generation efficiency.}
\end{figure*}

We present the analogy of the solution speed for all four arities and the fixed default debug instruction in Figure~\ref{fig:seidr:epg-distribution}. 
In detail, we show the distribution of EPG values in all experiments to explore how many candidate updates are generated before the solution is found.
We zoom in to the cases with solutions found with up to the first 10 program candidates in Figure~\ref{fig:seidr:epg-distrib-solved-10} and show the EPG distribution with the step of 100 candidates in Figure~\ref{fig:seidr:epg-distrib-solved-100}. 
In addition, we break down the results into each tree arity in Figure~\ref{fig:seidr:epg-bf}, showing the EPG on a heatmap scale and the TPR as a number between 0 and 1, or the signs ``+'' if a problem is solved and ``-'' if the final TPR is 0. 

Out of 100 experiments for each language, in 21--24\% of runs in Python and C++, the draft program is already the solution (EPG=0). 
For 31--33\% of the experiments, the solution is found after discarding 5 candidates. 
Around half of the experiments do not generate more than 100 programs. 
However, 5 problems are solved with more than 500 generated programs in Python and 1 problem in C++ (with $N=10$).
The results imply that the first steps in updating the draft program are crucial for solving the problem. 
The chances of solving the problem in later stages of the search, such as after 100 programs have been generated, are low.
This confirms our initial assumption in Section~\ref{sec:trade-off-settings} that 1000 programs are sufficient.

To briefly analyze Figure~\ref{fig:seidr:epg-bf}, we observe that some problems are solved in both languages, whereas some others --- only in Python. 
In addition, only five problems are not solved in any \method{} configuration in Python (bouncing-balls, bowling, cut-vector, dice-game and leaders) and seven in C++ (bouncing-balls, cut-vector, dice-game,  leaders, shopping-list, snow-day, and vector-distance).
Upon closer inspection of generated programs, we have noticed that in bouncing-balls, the programs have logical errors and differ considerably between programming languages, as well as in the majority of unsolved problems. 
Test cases and debug instructions in bowling frequently skewed the resulting programs to return answers to individual bowling score strings instead of writing an algorithm to calculate the score based on each next character.
The latter mistake happened in other unresolved problems, such as cut-vector.
Qualitative analysis has also shown that some programs failed to read input from the user and instead defined input strings within the code, which limited the program to testing only one I/O pair, although the algorithm was correct.

\begin{framed}\noindent
\textbf{Repair-replace trade-off in the initial exploration (\rqtreearity{}):} 
\method{} with Codex as the coding LLM outperforms the PushGP baseline on PSB2 in Python and performs on par with it in C++ experiments with tree arity of 10. 
Search strategies with tree arity larger than one benefit from the replace possibility of the \method{} framework as a consequence of using variable temperature for Codex-edit.
The repair component is also crucial for the framework because the replace-only search policy (with tree arity of $\infty$) performs worse than the policies that alternate between replace and repair during the program update (with tree arity of 10 or 100).  
\end{framed} 

\subsection{Generalizability Experiments}

In this section, we present and discuss replace-repair trade-off results obtained with GPT-3.5 and Llama 3 and the effect of switching from tournament selection to lexicase selection with the best hyperparameter settings found for the trade-off. 
We count the number of fully solved problems (i.e., reached $TPR=1$) in experiments with the hyperparameter settings described in Table~\ref{tab:w-n-generalizability} in six runs and present the language-specific results for PSB2 and HumanEval.
We also explore the total number of programs that need to be generated before a solution is obtained, as well as
in how many of the six runs each problem is fully solved.
The figures are described in detail in the dedicated sections. 

\subsubsection{Repair-replace Trade-off and Robustness to Restarts in the Generalizability Experiments.}
\label{sec:treearity-ollama}\label{sec:results-rqllama}

\begin{figure}[bt]
\begin{subfigure}{\linewidth}
\centering
\includegraphics[width=\linewidth, trim={0mm 0mm 0mm 0mm}]{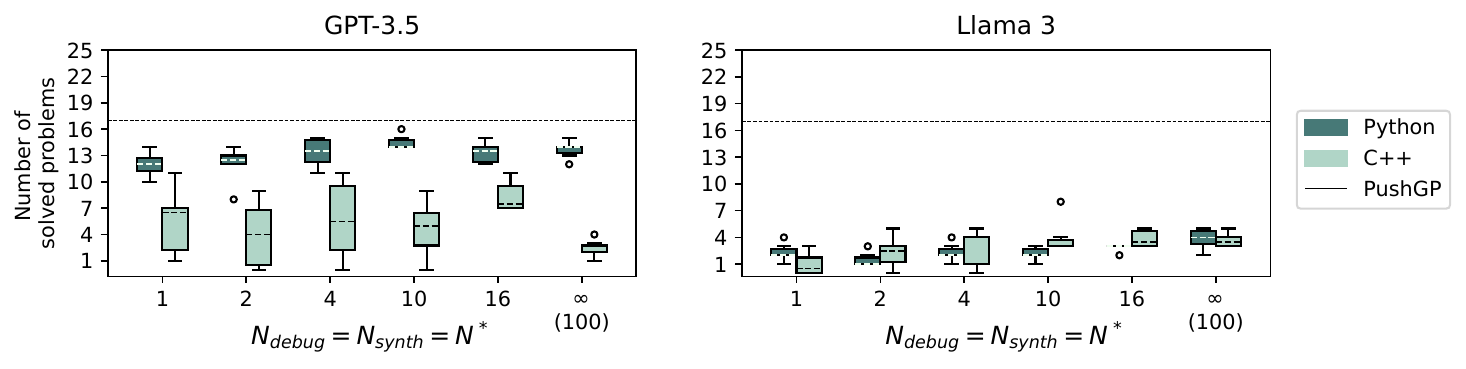}
  \caption{PSB2}
  \label{fig:num-solved-psb2-gpt3.5}
\end{subfigure}
\begin{subfigure}{\columnwidth}
\centering
\includegraphics[width=\linewidth, trim={0mm 0mm 0mm 0mm}]{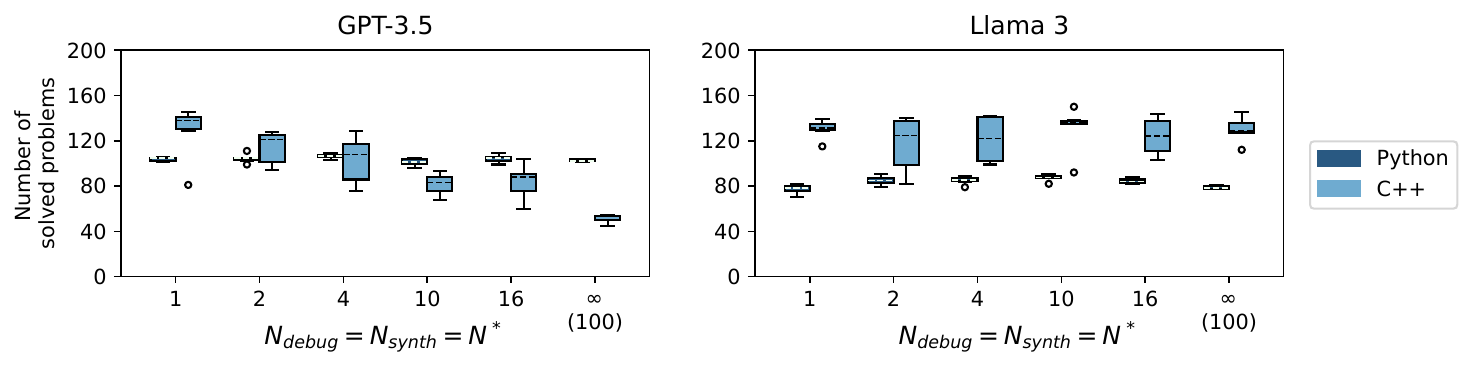}
  \caption{HumanEval}
  \label{fig:num-solved-he-gpt3.5}
\end{subfigure}
\caption{Repair-replace trade-off as a tree search problem in \method{}: the total number of solved problems as measured by $TPR=1$ using \method{} with GPT-3.5 and Llama 3 depending on tree arity $N^*$.}
\Description{Impact of tree arity on solved problems in SEIDR using GPT-3.5 and Llama 3.
The figure presents box plots showing the number of solved problems for different tree arities (N*) in the SEIDR framework using GPT-3.5 and Llama 3. Results are shown separately for the PSB2 (top row) and HumanEval (bottom row) benchmarks. The x-axis represents the number of debugging and synthesis attempts (all set to N*), while the y-axis indicates the number of solved problems. Python and C++ results are shown in different shades, with PushGP performance represented as a dotted line for PSB2.}
\label{fig:repair-replace-trade-off-generalizability}
\end{figure}

We compare the number of solved problems in the experiments with $\treearitydraft{}=\treearitydebug{}$ values of 1, 2, 4, 10, 16, $\infty$ (100) and $\treearityexplain{}=2$ in Python and C++ and tournament selection ranking strategy in Figure~\ref{fig:repair-replace-trade-off-generalizability}. 
We will refer to the equally set values of $\treearitydraft{}$ and $\treearitydebug{}$ as $N^*$ hereafter.
As before, experiments with $N^*=1$ and $N^*=\infty$ correspond to ablation studies, where the replace option or the repair option is turned off. 
The results of \method{} are compared to the baseline performance of PushGP on the PSB2 benchmark, which solves 17 out of 25 problems. 
The boxplots show the inter-quartile range between the first and third quartiles observed in six runs and the median values as horizontal lines within the boxes. 

The trend present for all the datasets and models is that the results for Python are more condensed over six runs than for C++. 
Since access to statistics about the training data for Llama 3 and GPT-3.5 is not provided, reasons for more stable performance in Python across runs can possibly lie in the training data distribution but cannot be confirmed. 
However, the fact that HumanEval-Python is a popular code generation benchmark against which models are compared may affect the results on this and other Python benchmarks.
In the same line of comparison of results between two programming languages, \method{} with \gpt{} performs better in \py{} than in \cpp{} on PSB2. 

\llama{} performs worse than \gpt{} on PSB2 in both languages.
This dataset has more test cases in stock than HumanEval and can be considered a more thorough test of the coding and debugging capabilities of LLMs. 
Following a recent trend, where larger models outperform smaller ones, the results of \method{} on PSB2 confirm that the smaller model (\llama{}) performs worse than the larger one (\gpt{}).

Looking back at the initial experiments with Codex and PSB2, we notice the degradation of performance from Codex to GPT-3.5: \method{} with Codex solved 19 problems in Python and 17 in C++ with tree arity 10, while the best-performing result of \method{} GPT-3.5 is 16 (tree arity of 10, too) in Python and 11 in C++ (several tree arities, but not 10). 
This result can be explained by the focus of LLM builders on the generalization of knowledge and performance on a variety of tasks, while Codex specializes in code generation.
Moreover, due to increased costs from Codex to GPT-3.5, we decrease the maximum number of generated program candidates from 1000 for Codex to 100 for GPT-3.5. 
However, only two problems are solved with Codex with $EPG > 100$: indices of substring (at program candidate \#510) and substitution cipher (at candidate \#210) in Python, bowling (candidate \#912) and indices of substring (\#410) in C++. 
Meanwhile, some problems are solved by Codex earlier than at the debugging attempt \#100 and not solved by GPT-3.5 and vice versa. 
Therefore, the reduction of the maximum generated programs has only a partial effect on the difference between the results of the two models. 

\begin{figure}[bt]
\begin{subfigure}{\linewidth}
\centering
\includegraphics[width=\linewidth, trim={0mm 0mm 0mm 0mm}]{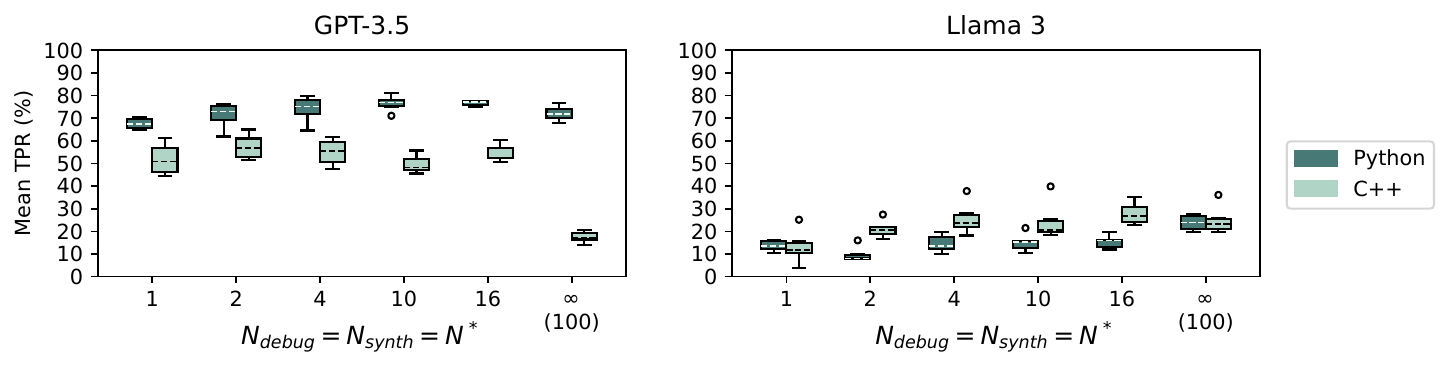}
  \caption{PSB2}
  \label{fig:mean-tpr-psb2-gpt3.5}
\end{subfigure}
\begin{subfigure}{\columnwidth}
\centering
\includegraphics[width=\linewidth, trim={0mm 0mm 0mm 0mm}]{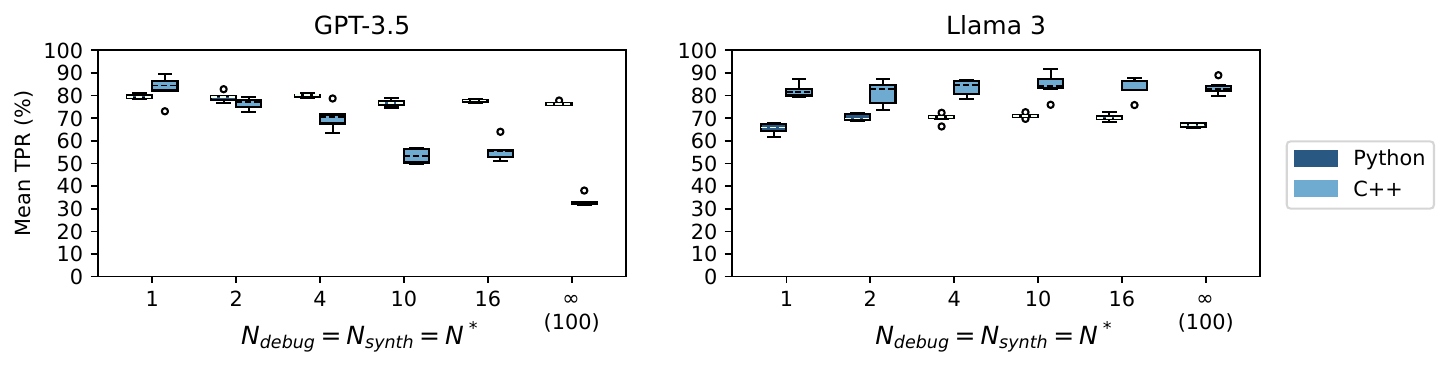}
  \caption{HumanEval}
  \label{fig:mean-tpr-he-gpt3.5}
\end{subfigure}
\caption{Repair-replace trade-off as a tree search problem in \method{}: mean $TPR$ measured in \% obtained using \method{} with GPT-3.5 and Llama 3 depending on tree arity $N^*$.}
\label{fig:mean-tpr-repair-replace-trade-off-generalizability}
\Description{Mean test pass rate (TPR) in SEIDR across different tree arities.
The figure presents box plots of the mean test pass rate (TPR) in percentage for different tree arities (N*), using GPT-3.5 and Llama 3. The results are shown for the PSB2 (top row) and HumanEval (bottom row) benchmarks. The x-axis represents the number of debugging and synthesis attempts (N*), while the y-axis indicates the mean TPR. Python and C++ results are displayed separately.}
\end{figure}

In addition to reporting the number of fully solved problems, Figure~\ref{fig:mean-tpr-repair-replace-trade-off-generalizability} also reports the mean Test Pass Rate measured in \%. 
For HumanEval-C++, \method{} with \gpt{} has better results with smaller tree arity values than with larger ones.
In Python, experiments with $1 < N^* < 100$ (i.e., non-corner case values of $N^*$) yield better mean TPR for \method{} with \gpt{} on PSB2 and with \llama{} --- for HumanEval-Python.
\method{} with \llama{} performs with slightly better mean TPR towards larger $N^*$  on PSB2-C++, and at the same time, it performs well on HumanEval-C++, so the difference between results with different $N^*$ is small.
The maximum of mean TPR for both datasets and models and max number of solved problems are obtained with $N^*=16$ or less, except for Llama 3 in PSB2-C++. 
From this part of the experiments, we notice that moderate or small tree arities $(N^* \le 16)$ are preferred, but there is no one leading tree arity.

\begin{figure}[tb]
  \centering
  \includegraphics[width=\linewidth, trim={3mm 2.8mm 3mm 2mm}, clip]{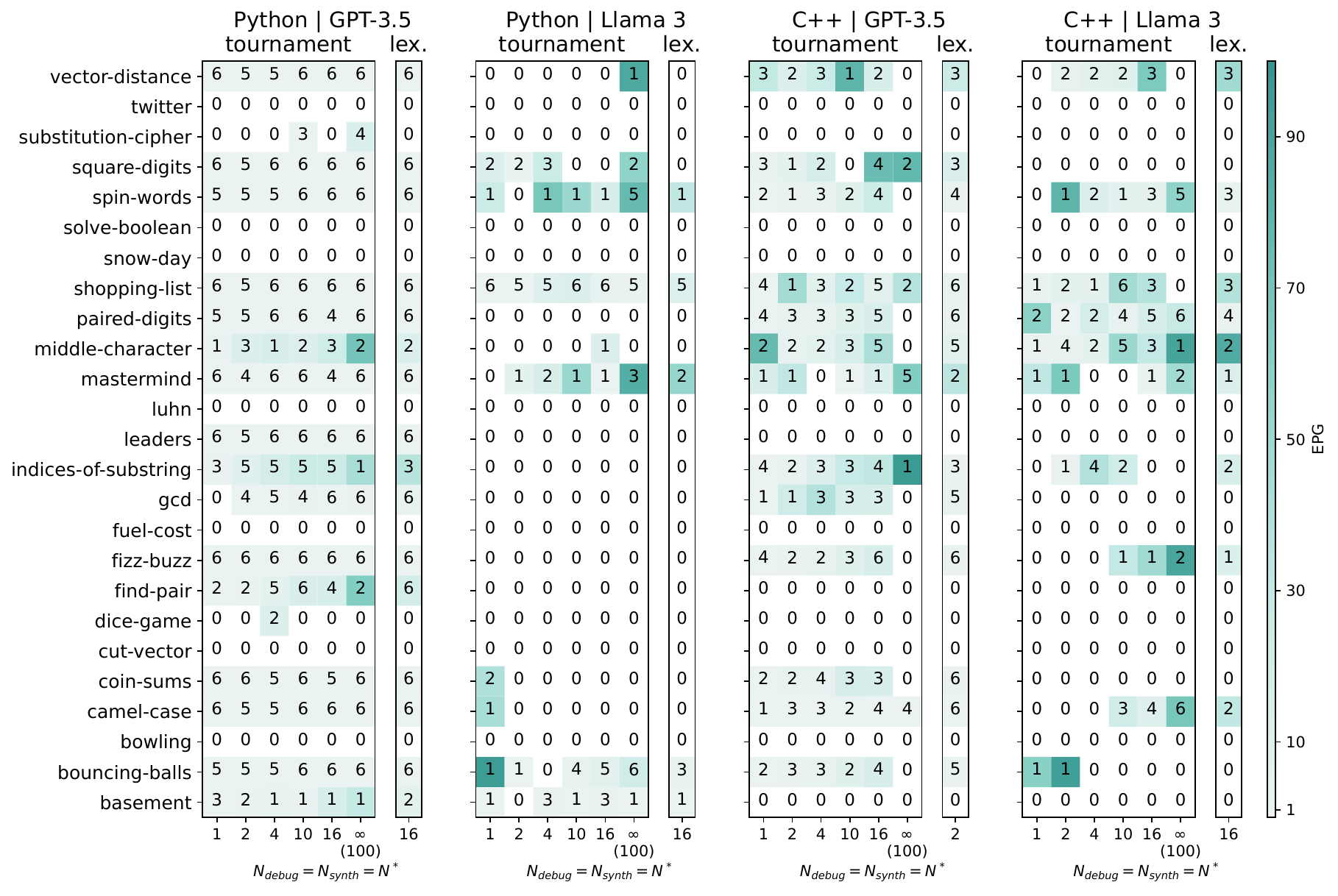}
  \caption{PSB2: mean Excess Programs Generated (in color) and the number of runs in which a task is solved. The experiments in which a specific problem is not solved in any run are shown in white.}
  \Description{Mean excess programs generated and solution frequency for PSB2 tasks.
The figure presents a heatmap comparing the mean number of excess programs generated (EPG) and the number of runs in which a task is solved for different configurations of SEIDR. The four panels correspond to Python and C++ using GPT-3.5 and Llama 3 under tournament and lexicase selection strategies. The x-axis represents the number of debugging and synthesis attempts (N*), while the y-axis lists the PSB2 benchmark tasks. Darker shades indicate higher EPG values, while the numbers in each cell represent the number of successful runs. White cells indicate tasks that were never solved in any run. The results highlight differences in problem-solving efficiency across programming languages, models, and selection strategies, with some configurations leading to more frequent solutions while others struggle with certain tasks.}
  \label{fig:epg-num-solved-psb2}
\end{figure}

\begin{figure}[hbt!]
  \centering
  \includegraphics[width=.88\linewidth, trim={0mm 3mm 0mm 2.6mm}, clip]{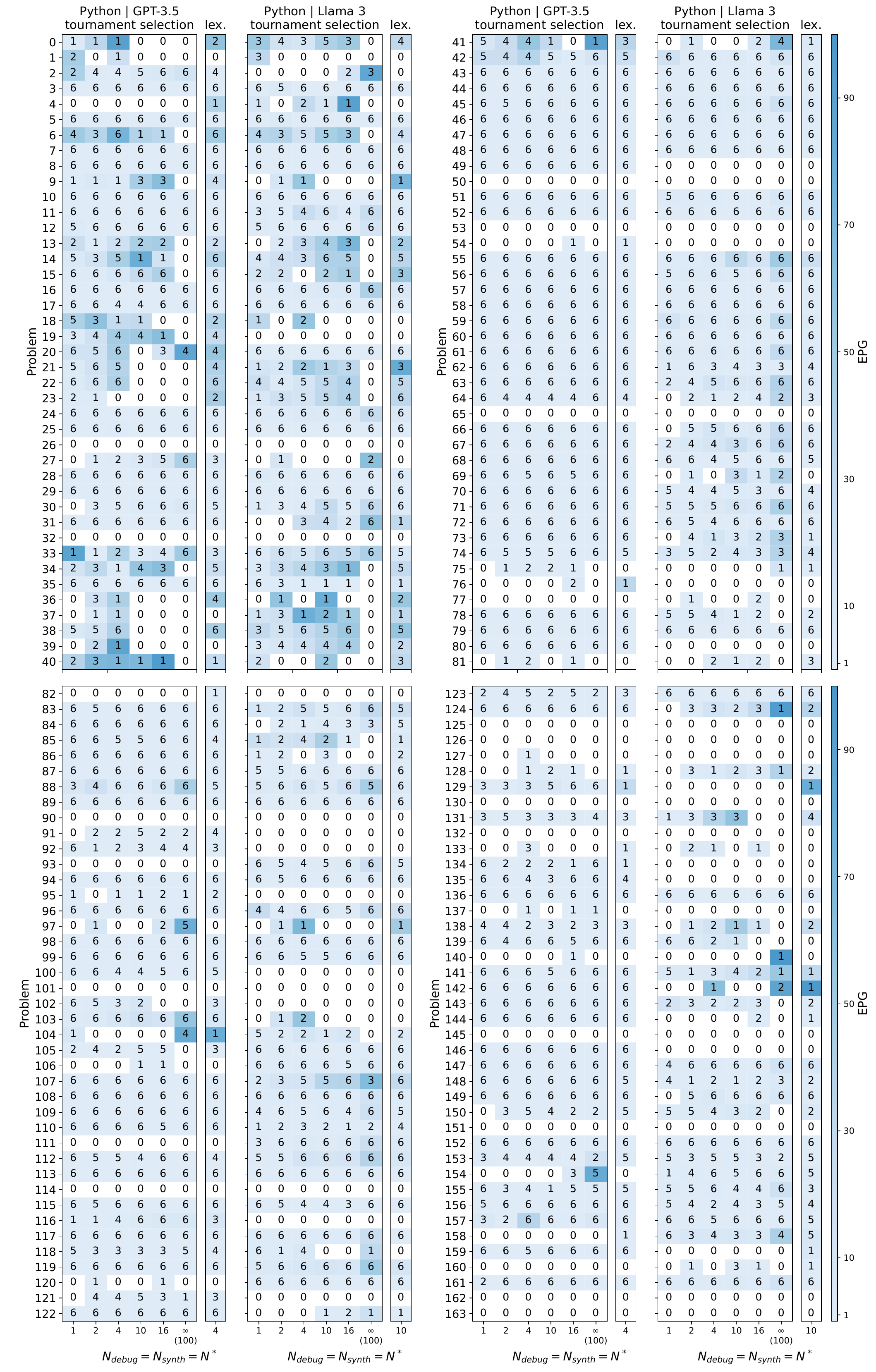}
  \vspace{-4pt}
  \caption{HumanEval-Python: mean Excess Programs Generated (in color) and the number of runs in which a task is solved. The problems that are not solved in any run are colored white.}
  \label{fig:epg-num-solved-he-python}
  \Description{Mean excess programs generated and solution frequency for HumanEval-Python.
The figure presents a heatmap showing the mean number of excess programs generated (EPG) and the number of runs in which a task is solved for the HumanEval benchmark using Python. The four panels compare results using GPT-3.5 and Llama 3 under tournament and lexicase selection strategies. The x-axis represents the number of debugging and synthesis attempts (N*), while the y-axis lists the HumanEval problems. Darker shades indicate higher EPG values, with numbers representing the number of successful runs. White cells indicate problems that were not solved in any run. The results illustrate differences in the effectiveness of various search and selection strategies, revealing distinct patterns in problem difficulty and program generation efficiency. Some problems remain unsolved regardless of the model or selection method.}
  \vspace{-12pt}
\end{figure}

\begin{figure}[hbt!]
  \centering
  \includegraphics[width=.88\linewidth, trim={0mm 3mm 0mm 2.6mm}, clip]{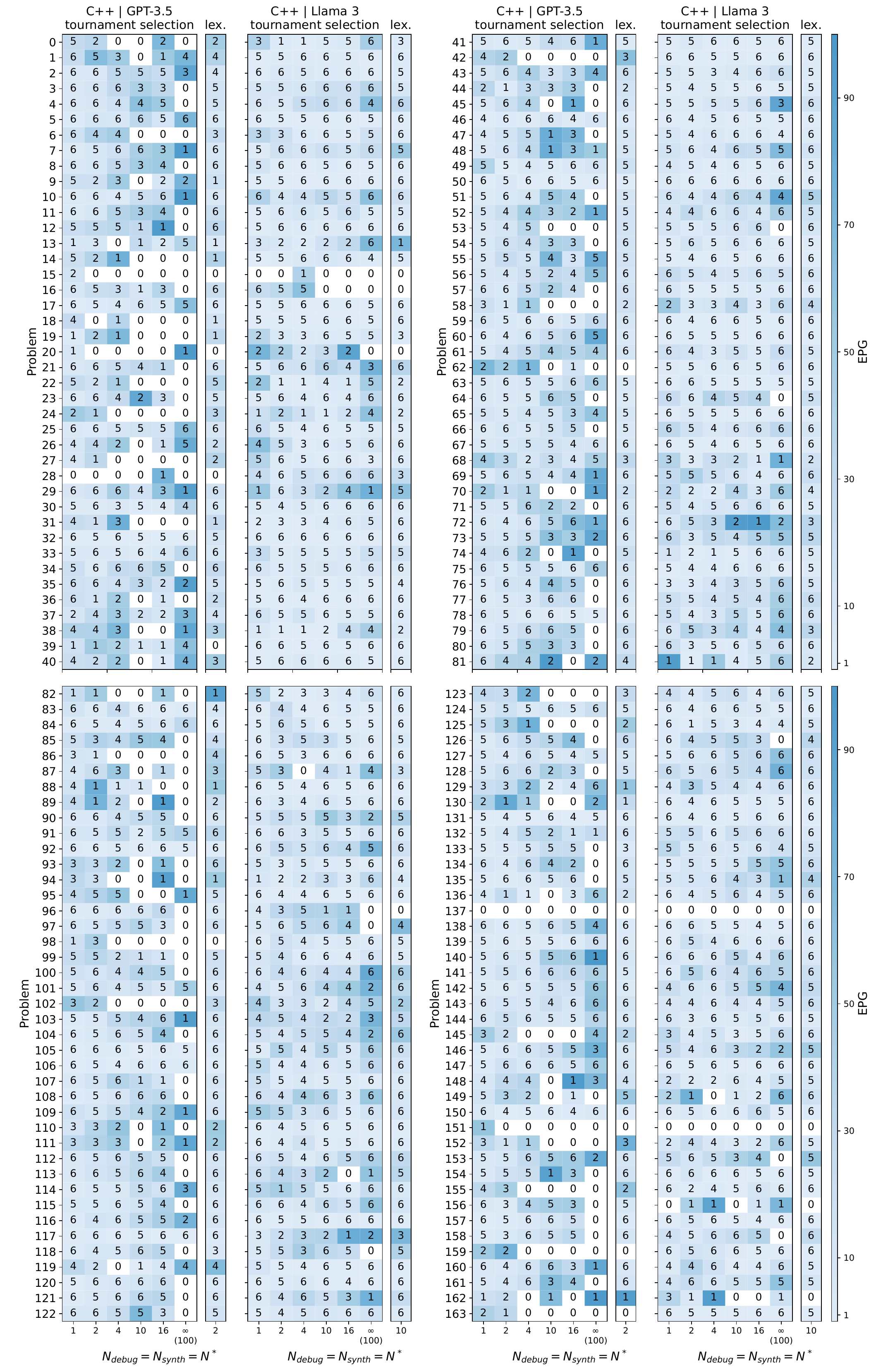}  %
  \vspace{-4pt}
  \caption{HumanEval-C++: mean Excess Programs Generated (in color) and the number of runs in which a task is solved. The problems that are not solved in any run are colored white.}
  \label{fig:epg-num-solved-he-c++}
  \Description{Mean excess programs generated and solution frequency for HumanEval-C++.
The figure is a complement to the previous one but uses C++ instead of Python. It presents a heatmap showing the mean number of excess programs generated (EPG) and the number of runs in which a task is solved for the HumanEval benchmark using C++. The four panels compare results using GPT-3.5 and Llama 3 under tournament and lexicase selection strategies. The x-axis represents the number of debugging and synthesis attempts (N*), while the y-axis lists the HumanEval problems. Darker shades indicate higher EPG values, while numbers in each cell represent the number of successful runs. White cells indicate tasks that were not solved in any run. 
The results illustrate differences in the effectiveness of various search and selection strategies, revealing distinct patterns in problem difficulty and program generation efficiency. Some problems remain unsolved regardless of the model or selection method.}
    \vspace{-12pt}
\end{figure}

The number of runs in which a problem is solved and the average speed of finding those solutions are shown in Figure~\ref{fig:epg-num-solved-psb2} for PSB2, Figure~\ref{fig:epg-num-solved-he-python}
for HumanEval-Python and Figure~\ref{fig:epg-num-solved-he-c++} for HumanEval-C++.
These figures show which problems were not solved in any run of any experiment (a row with zeros colored white) and what problems are easier to solve than others (e.g., solved in all runs, or at least one with each tree arity $N^*,$ or earlier in the search tree and shown in brighter rather than darker color but not white). 
We also show the results of lexicase selection runs marked with ``lex.'' in these three figures, but will discuss them in a separate section. 

The majority of solved PSB2 problems are solved in more than one run per setting with \gpt{} and faster (with fewer attempts) in Python than in C++ as illustrated in Figure~\ref{fig:epg-num-solved-psb2}.
For \llama{}, most of the solutions are obtained in 1--3 runs.
The trend for both datasets is that, for problems where a solution is found, \method{} with \gpt{} makes fewer attempts in Python (brighter colors prevail in the corresponding parts of Figures~\ref{fig:epg-num-solved-psb2} and~\ref{fig:epg-num-solved-he-python}) than \method{} with \gpt{} in C++ or \method{} with \llama{} in both languages (darker shades in the \gpt{} on C++ and \llama{} parts of Figures~\ref{fig:epg-num-solved-psb2} and \ref{fig:epg-num-solved-he-c++}).

To explore the capabilities of \method{} with the studied models, we can take a union over experiments and count the number of problems solved at least once across 36\footnote{~We have 6 runs and 6 different values of $N^*.$} restarts with a fixed dataset, language and model.
In this way, in addition to reporting the number of problems solved in each run in the boxplot, we can derive the number of problems solved in any run with any set of hyperparameter using information in Figures~\ref{fig:epg-num-solved-psb2}--\ref{fig:epg-num-solved-he-c++}.
For example, \method{} with \gpt{} does not solve only 7 problems out of 25 PSB2-Python tasks (see Figure~\ref{fig:epg-num-solved-psb2}).
Namely, bouncing-balls, cut-vector, dice-game, indices-of-substring, middle-character, solve-boolean, and substitution-cipher are solved in 0 runs with all the variations of $N^*$.
In other words, 18 unique problems are solved in the collection of all the experiments using \method{} with \gpt{} on PSB2-Python. 
\method{} with \gpt{} does not solve 12 out of 25 PSB2-C++ tasks with any hyperparameter settings. 
The number of solved problems in any run with \method{} and \llama{} are 10 for PSB2 in both languages. 
To support the finding about degradation of performance happening in a non-specialized model GPT-3.5 compared to the code-specialized Codex, we calculate the number of solved problems in any run with Codex and obtain 20 problems solved at least once in Python and 18 in C++. 

Similarly, \method{} with \gpt{} solves 141 out of 164 HumanEval-Python problems at least once in all runs, collectively, and 128 problems with \llama{} (see Figure~\ref{fig:epg-num-solved-he-python}).
In HumanEval-C++, \method{} with \gpt{} solves 163 out of 164 problems (except CPP/137) in the union of all runs and 162 with \llama{} (except CPP/137 and CPP/151) as follows from Figure~\ref{fig:epg-num-solved-he-c++}.

\begin{figure}[tb]
\begin{subfigure}{.8\columnwidth}
\centering
\includegraphics[width=\linewidth, trim={0mm 3mm 0mm 0mm}, clip]{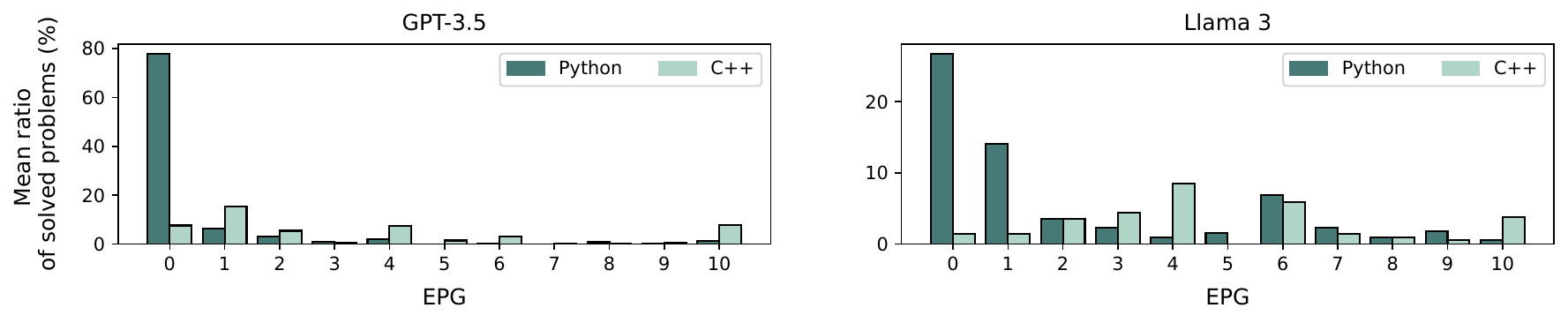}
  \caption{PSB2: 0 $\leq$ EPG $\leq$ 10 with step 1.}
  \label{fig:psb2-epg-distrib-step-1}
\end{subfigure}
\begin{subfigure}{.8\columnwidth}
\centering
\includegraphics[width=\linewidth, trim={0mm 3mm 0mm 0mm}, clip]{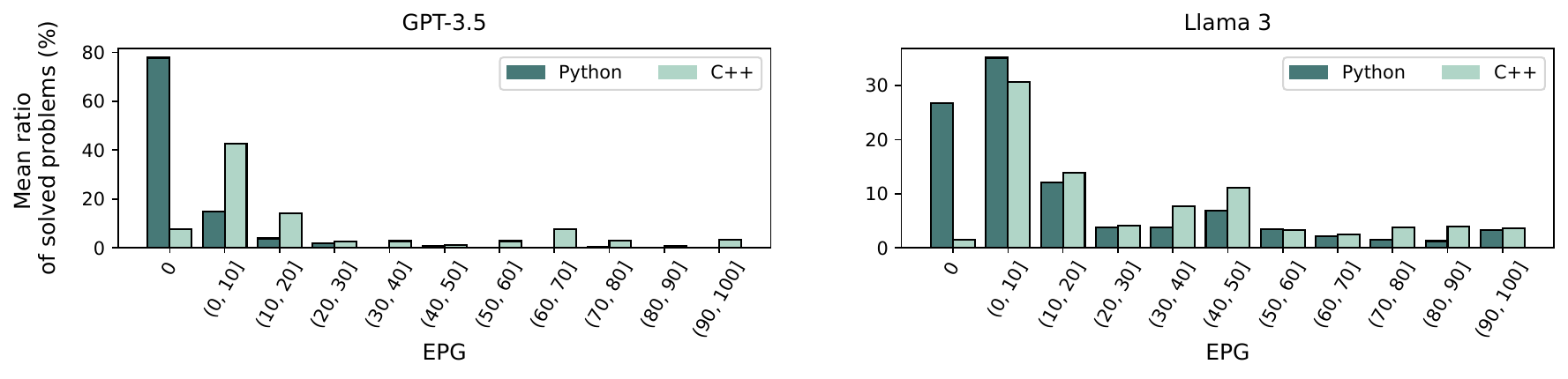}
  \caption{PSB2: 0 $\leq$ EPG $\leq$ 100 with step 10.}
  \label{fig:psb2-epg-distrib-step-10}
\end{subfigure}
\begin{subfigure}{.8\columnwidth}
\centering
\includegraphics[width=\linewidth, trim={0mm 3mm 0mm 0mm}, clip]{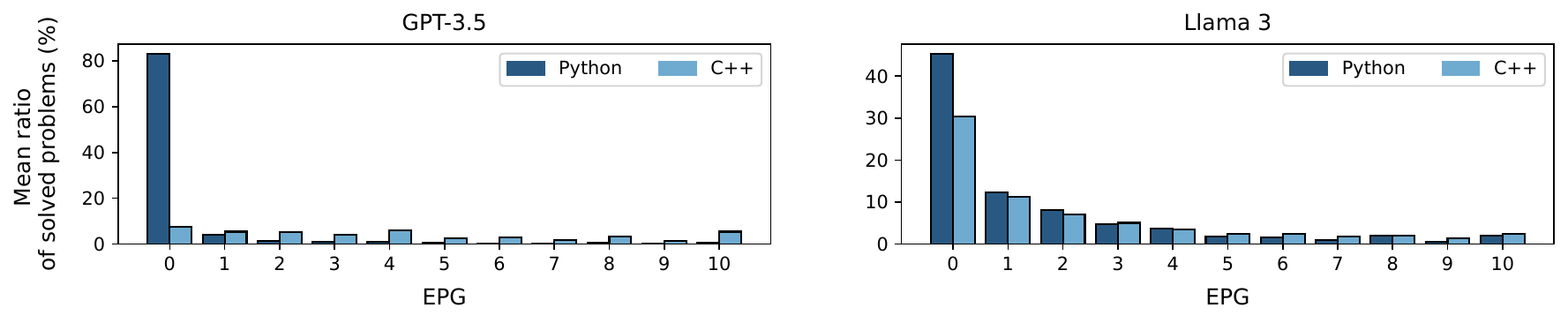}
  \caption{HumanEval: 0 $\leq$ EPG $\leq$ 10 with step 1.}
  \label{fig:humaneval-epg-distrib-step-1}
\end{subfigure}
\begin{subfigure}{.8\columnwidth}
\centering
\includegraphics[width=\linewidth, trim={0mm 3mm 0mm 0mm}, clip]{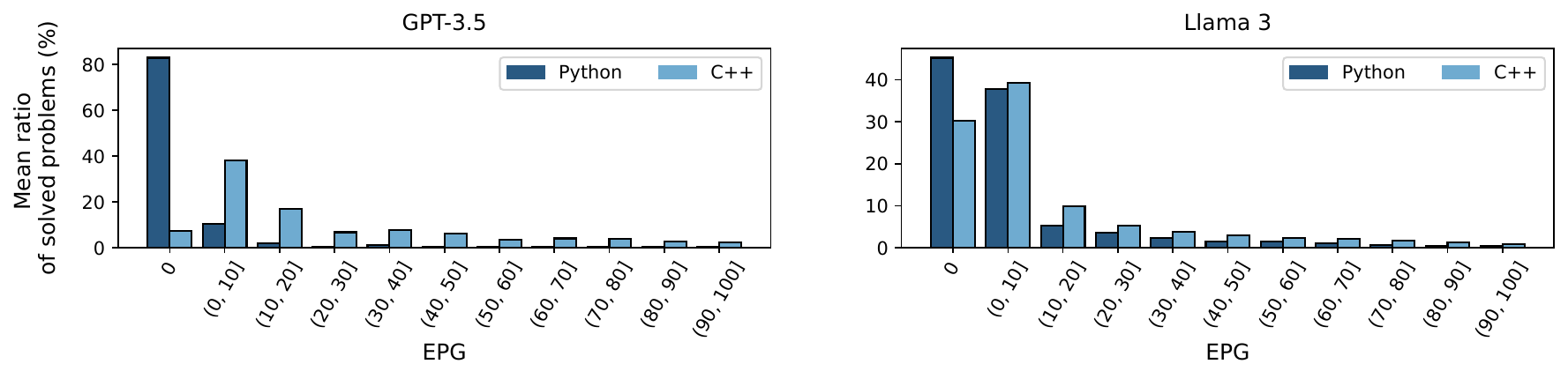}
  \caption{HumanEval: 0 $\leq$ EPG $\leq$ 100 with step 10.}
  \label{fig:humaneval-epg-distrib-step-10}
\end{subfigure}
\caption{Distribution (in \%) of the number of generated programs with GPT-3.5 and Llama 3 during each problem-solving attempt on average over 6 runs with different tree arities $\treearitydraft{}, \; \treearitydebug{}.$}
\label{fig:epg-distribution}
\Description{Distribution of generated programs during problem-solving attempts.
The figure presents histograms showing the percentage distribution of excess programs generated (EPG) during problem-solving attempts across different benchmarks and models. The x-axis represents EPG values, while the y-axis indicates the mean ratio of solved problems in percentage. The top two rows correspond to the PSB2 benchmark with different EPG ranges (0–10 with step 1 and 0–100 with step 10), while the bottom two rows show results for the HumanEval benchmark with the same EPG ranges. Results are shown separately for GPT-3.5 and Llama 3, with comparisons between Python and C++. The distributions reveal that most problems are solved with lower EPG values, particularly with GPT-3.5, while Llama 3 exhibits a wider spread, suggesting that more generated programs are required for successful problem-solving.}
\end{figure}

In Figure~\ref{fig:epg-distribution}, we present the analogy of the speed of obtaining solutions, EPG. 
We zoom in to the cases with solutions found with up to the first 10 program candidates in Figures~\ref{fig:psb2-epg-distrib-step-1} and~\ref{fig:humaneval-epg-distrib-step-1} for PSB2 and HumanEval-X, respectively. 
The coarser-grained EPG distribution with the step of 10 candidates is shown in Figures~\ref{fig:psb2-epg-distrib-step-10} and~\ref{fig:humaneval-epg-distrib-step-10}. 

In detail, we show the distribution of EPG values in all experiments to explore what proportion of candidate updates is made before a solution is found.
For example, on average, over 70\% of Python solutions by \method{} with \gpt{} are solved from the first attempt, i.e., have EPG$=0$ (see Figure~\ref{fig:psb2-epg-distrib-step-1}).
Python solutions are more frequently found from the first attempt by \method{} with \llama{} than later in the tree, although less frequently than with \gpt{}.
Most of solutions found by \method{} benefit from the iterative repair and generate up to 10 extra programs with both models (see Figures~\ref{fig:psb2-epg-distrib-step-10}, \ref{fig:humaneval-epg-distrib-step-10}). 
However, some solutions are also found later in the tree search. 

The EPG distribution results for correctly solved problems with $TPR=1$ imply that the first steps in the update of the draft program are crucial for solving the problem. 
The chances of solving the problem in the later stages of the search are low.
This confirms our assumption in Section~\ref{sec:trade-off-settings} that 100 programs are sufficient in the generalizability experiments.

\begin{framed}\noindent
\textbf{Repair-replace trade-off for \method{} with GPT-3.5 and \llama{} in the generalizability experiments (\rqtreearity{}, \rqllama{}, \rqmultirun{}):}
Unlike for \method{} with Codex and a fixed debugging instruction (see Section~\ref{sec:seidr:rqtreearity}), \method{} with \gpt{} and \llama{} does not show any distinct trend for all the languages and datasets in terms of the preferred tree arity value. 
Results over a number of runs with the same hyperparameter settings are more condensed for Python than for C++, which can be a result of optimizing LLMs for high performance on popular coding benchmarks in Python.
If solutions to problems are found by \method{}, it is done at an earlier tree search step for Python than for C++ (i.e., with a smaller EPG). 
The majority of solutions are found within the first 10 updates of program candidates. 
\method{} solves 163 out of 164 HumanEval-C++ problems with \gpt{} at least once over all runs with all restarts and hyperparameter sets, and 162 --- with \llama{}.
\end{framed}

\subsubsection{Parent Selection Strategies within the Ranking Agent in the Generalizability Experiments}
\label{sec:lexicase-results}

\begin{figure}[bt]
\begin{subfigure}{.45\linewidth}
\centering
\includegraphics[width=\linewidth, trim={0mm 4mm 0mm 0mm}]{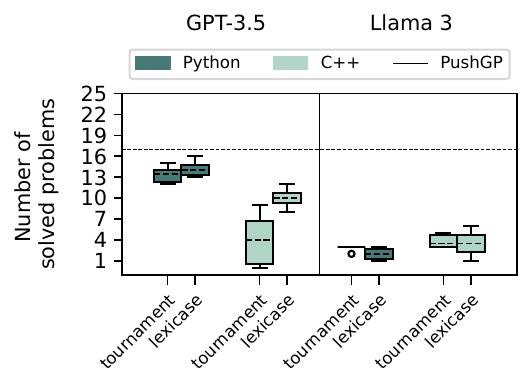}
  \caption{PSB2.}
  \label{fig:num-solved-lexicase-selection-psb2}
\end{subfigure}
\begin{subfigure}{.45\linewidth}
\centering
\includegraphics[width=\linewidth, trim={0mm 4mm 0mm 0mm}]{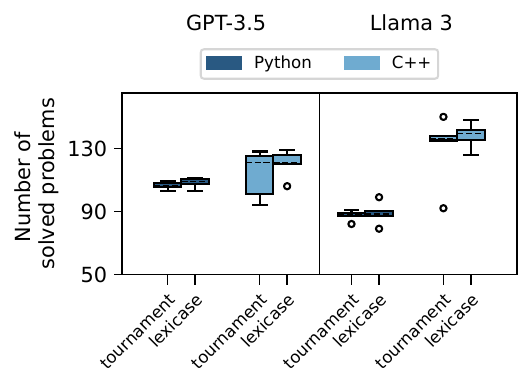}
  \caption{HumanEval.}
  \label{fig:num-solved-lexicase-selection-he}
\end{subfigure}
\caption{Number of solved problems in lexicase and tournament selection experiments for \method{} with GPT-3.5 and Llama 3 and fixed tree arity $N^*$.}
\label{fig:num-solved-lexicase-selection}
\Description{Comparison of solved problems using lexicase and tournament selection.
The figure presents box plots showing the number of problems solved in SEIDR using GPT-3.5 and Llama 3 across two different selection strategies: tournament and lexicase selection. The left panel (a) corresponds to the PSB2 benchmark, while the right panel (b) corresponds to the HumanEval benchmark. Python and C++ results are displayed separately, with a dashed line indicating PushGP performance for PSB2.}
\end{figure}

\begin{figure}[bt]
\begin{subfigure}{.45\linewidth}
\centering
\includegraphics[width=\linewidth, trim={0mm 4mm 0mm 0mm}]{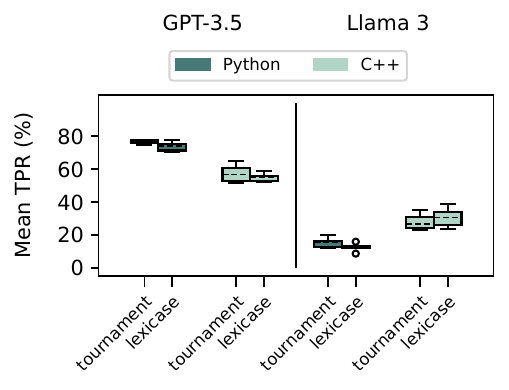}
  \caption{PSB2.}
  \label{fig:mean-tpr-lexicase-selection-psb2}
\end{subfigure}
\begin{subfigure}{.45\linewidth}
\centering
\includegraphics[width=\linewidth, trim={0mm 4mm 0mm 0mm}]{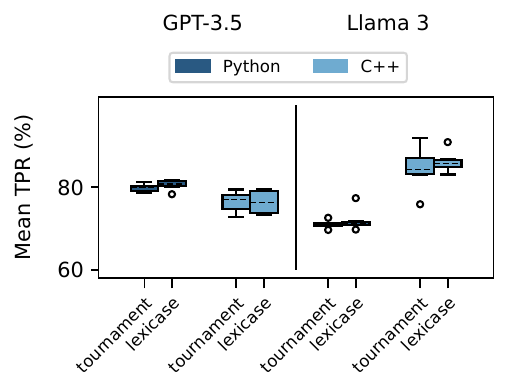}
  \caption{HumanEval.}
  \label{fig:mean-tpr-lexicase-selection-he}
\end{subfigure}
\caption{Mean $TPR$ (measured in \%) in lexicase and tournament selection experiments for \method{} with GPT-3.5 and Llama 3 and fixed tree arity $N^*.$}
\label{fig:mean-tpr-lexicase-selection}
\Description{Mean test pass rate (TPR) for lexicase and tournament selection.
The figure presents box plots of the mean test pass rate (TPR) in percentage for SEIDR experiments using GPT-3.5 and Llama 3 with lexicase and tournament selection. The left panel (a) corresponds to the PSB2 benchmark, while the right panel (b) corresponds to HumanEval. Python and C++ results are shown separately.}
\end{figure}

Based on the mean TPR over all runs reported in Figure~\ref{fig:mean-tpr-repair-replace-trade-off-generalizability}, we fix the best-performing $N^*$ for each dataset, language and model and run the experiments with lexicase selection instead of tournament selection as a parent selection strategy. 
The hyperparameters are shown in Table~\ref{tab:lexicase-selection-hyperparameters}.
We compare the number of problems solved with these settings and two types of parent selection algorithms in Figure~\ref{fig:num-solved-lexicase-selection} and mean TPR in Figure~\ref{fig:mean-tpr-lexicase-selection}.
Mean TPR with lexicase selection and the best tournament selection configuration are similar for two selection strategies for most of the experiments.
Similar results are obtained for the number of fully solved problems, with the exception of C++ results of \method{} with \gpt{} on both datasets, where lexicase selection improved the results.

\begin{table}[t]
\setlength{\tabcolsep}{4pt}
\centering
\caption{\method{} generalizability experiments: hyperparameters in the lexicase selection experiments.}\small
\label{tab:lexicase-selection-hyperparameters}
\begin{tabular}{rcccc|cccc}
\toprule
experiment \# & 22 & 23 & 24 & 25 & 26 & 27 & 28 & 29 \\
\midrule
datasets  & \multicolumn{4}{c|}{PSB2} & \multicolumn{4}{c}{HumanEval}  \\ 
\midrule
models  & 
\multicolumn{2}{c|}{GPT-3.5} &
\multicolumn{2}{c|}{Llama 3} &
\multicolumn{2}{c|}{GPT-3.5} &
\multicolumn{2}{c}{Llama 3}\\ 
\midrule
language  & C++ & Python & C++ & \multicolumn{1}{c|}{Python} & C++ & Python & C++ & Python \\
\midrule
population size (beam width), $\beamwidth{}$ & 2 & 16 & 16 & 16 & 2 & 4 & 10 & 10  \\
\# programs in the 1st generation, $\treearitydraft{}$ & 2 & 16 & 16 & 16 & 2 & 4 & 10 & 10  \\
\# bug explanations for candidate, $\treearityexplain{}$ & 2 & 2 & 2 & 2 & 2 & 2 & 2 & 2 \\
\# repairs for each explanation, $\treearitydebug{}$ & 2 & 16 & 16 & 16 & 2 & 4 & 10 & 10 \\
\midrule
max programs generated & \multicolumn{8}{c}{100} \\
prompts & \multicolumn{8}{c}{see Section~\ref{sec:ollama-prompts}} \\
\# runs per experiment &  \multicolumn{8}{c}{6} \\
\bottomrule
\end{tabular}
\end{table}

To zoom in on the improvement details, we refer back to Figures~\ref{fig:epg-num-solved-psb2}--\ref{fig:epg-num-solved-he-c++}, where separate columns are dedicated to lexicase selection experiments (marked with ``lex.'').
We do not observe different programs solved with lexicase selection than with tournament selection. 
Some problems are solved in all six runs with lexicase selection and in fewer runs with tournament selection, such as find-pair and fizz-buzz in the C++ experiments with \gpt{}.

\begin{figure}[bt]
\centering
\includegraphics[width=\linewidth, trim={2mm 123mm 3mm 130mm}, clip]{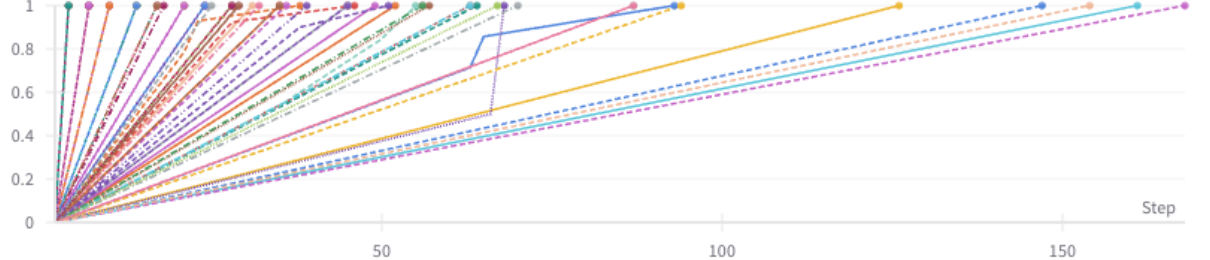}
\caption{The best score on validation tests (y-axis) obtained up to an indicated logging step (x-axis) for problems solved from the second or later attempts with lexicase selection using \method{} with \gpt{}. Results are obtained for HumanEval. ``Step'' on the x-axis corresponds to logging settings and roughly indicates the speed of finding solutions to different problems. Each line stands for a unique solution.}
\label{fig:lexicase-tpr-jumps}
\Description{Progression of validation test scores over logging steps in SEIDR.
The figure shows the best validation test scores (y-axis) obtained at different logging steps (x-axis) for HumanEval problems solved from the second or later attempts using lexicase selection in SEIDR with GPT-3.5. Each line represents a unique solution, illustrating how different problems reach their best validation scores over time. The x-axis, labeled “Step,” corresponds to logging intervals and provides an indication of the speed at which solutions are found. The trajectories of the lines reveal variations in convergence rates, with some problems achieving high scores rapidly while others require more steps to reach an optimal solution.}
\end{figure}

To confirm the effect of lexicase selection on the program candidate search, we have explored the test pass rate dynamics.
An example run of \method{} with \gpt{} on HumanEval is shown in Figure~\ref{fig:lexicase-tpr-jumps}.
We observe that in the vast majority of cases, the test score jumps from 0 to 1 directly, not as a result of reordering candidates, but as a result of \method{}  bug summarization and candidate update.
Therefore, we appoint differences in results between lexicase and tournament parent selection primarily to the LLMs themselves and the debugging loop rather than to the parent selection strategy. 
Specifically, when an LLM gets the prompt that fixes exactly the error in a program candidate, the score jumps to 1 because of the correct prompt provided to the model more frequently than because a better candidate is chosen on a previous step.

\begin{framed}\noindent
\textbf{Parent selection strategies for \method{} with GPT-3.5 and \llama{} in the generalizability experiments (\rqllama{}, \rqmultirun{}, \rqlexicase{}):} 
No leading ranking strategy is found for \method{} experiments with GPT-3.5 and \llama{} as measured by the average test pass rate and the average number of solved problems over six runs. 
In the vast majority of experiments where a solution is found and at least one debugging attempt is made the score jumps from 0 to 1 as opposed to climbing up incrementally and makes the ranking strategy less impactful than effective prompting. 
\end{framed}

\begin{table}[t]
    \centering
    \caption{Average number of solved PSB2 problems using \method{} with different LLMs and PushGP. The best results among \method{}-only experiments at each $k$ are highlighted in bold. If pass@100 is not available, we cite the best available results in brackets. $N^*$ stands for the number of debug candidates and debug instructions to generate (tree arity).}\small
    \label{tab:generalizability-psb2}
\begin{adjustbox}{max width=.92\textwidth}
\begin{DIFnomarkup} %
\begin{tabular}{llllrrr}
\toprule
Language & Model in \method{} & $N^*$ & Parent selection &  pass@1 &  pass@10 &  pass@100 \\
\midrule
Python & \gpt{} & 1   &         tournament &    10.5 &     12.0 &      12.0 \\
    &        & 2   &         tournament &     9.3 &     11.5 &      12.0 \\
    &        & 4   &         tournament &     9.8 &     12.2 &      13.3 \\
    &        & 10  &         tournament &    10.7 &     \textbf{12.7} &      \textbf{14.5} \\
    &        & 16  &         tournament &    10.0 &     11.3 &      13.3 \\
    &        & 16  &           lexicase &    10.2 &     12.3 &      14.2 \\
    &        & 100 &         tournament &    10.8 &     12.2 &      13.7 \\
\cline{3-7}\\[-8pt]
    &  \multicolumn{3}{l}{solved at least once by \method{} with \gpt{}} &  13 &       17 &       18 \\[1pt]
\cline{2-7}\\[-8pt]
        & Codex  & 10 & tournament &      5 &       10 &        14 (pass@1000=19) \\[1pt]
\cline{3-7}\\[-8pt]
       &  \multicolumn{3}{l}{solved at least once by \method{} with Codex}   & 8 &       13 &        17 (pass@1000=20) \\[3pt]
\cline{2-7}\\[-8pt]
        &  PushGP (no \method{})   &    -            &       - &        - &       - &  (17)\\[1pt]
\cline{2-7}\\[-8pt]
        & \llama{} & 1   &         tournament &     1.2 &      1.3 &       2.3 \\
        &        & 2   &         tournament &     1.0 &      1.5 &       1.5 \\
        &        & 4   &         tournament &     1.0 &      1.5 &       2.3 \\
        &        & 10  &         tournament &     1.5 &      2.0 &       2.2 \\
        &        & 16  &         tournament &     1.7 &      \textbf{2.8} &       2.8 \\
        &        & 16  &           lexicase &     1.0 &      1.7 &       2.0 \\
        &        & 100 &         tournament &     1.0 &      1.0 &       \textbf{3.8} \\[1pt]
\cline{3-7}\\[-8pt]
    &  \multicolumn{3}{l}{solved at least once by \method{} with \llama{}} &   4 &        5 &       11 \\
\midrule\\[-8pt]
 C++ & GPT-3.5 & 1   &         tournament &     0.0 &      5.6 &       5.5 \\
       &        & 2   &         tournament &     1.0 &      5.0 &       6.0 \\
       &        & 2   &           lexicase &     1.3 &     \textbf{ 9.0} &      \textbf{10.0} \\
       &        & 4   &         tournament &     1.0 &      4.6 &       6.8 \\
       &        & 10  &         tournament &     1.0 &      1.3 &       5.6 \\
       &        & 16  &         tournament &     1.0 &      1.2 &       8.3 \\
       &        & 100 &         tournament &     1.0 &      1.2 &       2.3 \\[1pt]
\cline{3-7}\\[-8pt]
       & \multicolumn{3}{l}{solved at least once by \method{} with \gpt{}}   & 2 &       13 &       13 \\[1pt]
\cline{2-7}\\[-8pt]
   & Codex  & 10 & tournament &     3 &       12 &   14 (pass@1000=17)\\[1pt]
\cline{3-7}\\[-8pt]
       & \multicolumn{3}{l}{solved at least once by \method{} with Codex}   & 10 &       12 &        15 (pass@1000=18) \\[1pt]
\cline{2-7}\\[-8pt]
   &  PushGP (no \method{}) & -   &    -            &       - &        - &      (17)\\[1pt]
\cline{2-7}\\[-8pt]
        & Llama 3 & 1   &         tournament &     0.0 &      1.0 &       2.0 \\
       &        & 2   &         tournament &     0.0 &      \textbf{2.0} &       2.8 \\
       &        & 4   &         tournament &     1.0 &     \textbf{ 2.0} &       2.6 \\
       &        & 10  &         tournament &     1.0 &      1.4 &       \textbf{4.0} \\
       &        & 16  &         tournament &     0.0 &      1.8 &       3.8 \\
       &        & 16  &           lexicase &    1.0 &      1.4 &       3.5 \\
       &        & 100 &         tournament &     0.0 &      0.0 &       3.7 \\[1pt]
\cline{3-7}\\[-8pt]
       & \multicolumn{3}{l}{solved at least once by \method{} with \llama{}}  & 4 &        5 &       11  \\
\bottomrule
\\ %
\\ %
\end{tabular}
\end{DIFnomarkup} %
\end{adjustbox}
\end{table}

\begin{table}[t]
    \centering
    \caption{Percentage of solved tasks in HumanEval-X using \method{} with different LLMs. The best results among \method{}-only experiments at each $k$ are highlighted in bold. $N^*$ stands for the number of debug candidates and debug instructions to generate (tree arity).}\small
    \label{tab:generalizability-he}
\begin{adjustbox}{max width=\textwidth}
\begin{DIFnomarkup} %
\begin{tabular}{llllrrr}
\toprule
Language & Model in \method{} & $N^*$ & Ranking &  pass@1 &  pass@10 &  pass@100 \\
\midrule
Python & \gpt{} & 1   &         tournament &    53.4 &     \textbf{60.4} &      63.3 \\
       &        & 2   &         tournament &    51.3 &     58.2 &      63.7 \\
       &        & 4   &         tournament &    52.1 &     58.4 &      64.8 \\
       &        & 4   &           lexicase &    52.5 &     59.6 &      \textbf{66.1} \\
       &        & 10  &         tournament &    51.1 &     58.3 &      61.7 \\
       &        & 16  &         tournament &    53.0 &     59.6 &      63.5 \\
       &        & 100 &         tournament &    54.0 &     56.7 &      62.5 \\[1pt]
\cline{3-7}\\[-8pt]
       & \multicolumn{3}{l}{solved at least once by \method{} with \gpt{}}   & 70.7 &     84.1 &      87.8 \\[1pt]
\cline{2-7}\\[-8pt]
    &   Llama 3 & 1   &         tournament &    22.0 &     44.0 &      47.1 \\
       &        & 2   &         tournament &    24.0 &     \textbf{48.5 }&      51.8 \\
       &        & 4   &         tournament &    22.5 &     43.1 &      51.8 \\
       &        & 10  &         tournament &    24.0 &     42.9 &      53.3 \\
       &        & 10  &           lexicase &    26.2 &     44.1 &      \textbf{54.1} \\
       &        & 16  &         tournament &    23.5 &     41.6 &      51.5 \\
       &        & 100 &         tournament &    21.4 &     26.3 &      48.0  \\[1pt]
\cline{3-7}\\[-8pt]
       & \multicolumn{3}{l}{solved at least once by \method{} with \llama{}} & 56.7 &     73.8 &      79.3 \\[1pt]
\cline{2-7}\\[-8pt]
& \multicolumn{6}{l}{\textbf{LLM results without \method{}}} \\
 & GPT-3.5 (ChatGPT) & - &  - &  48.1  &  -   &    - \\
 & GPT-4 & - &  - & 67.0   &  -   &    - \\
 & Code Llama 34B & - &  - &  48.8  &  76.8   &    93.0 \\
 & Unnatural Code Llama 34B & - &  - &  62.2  &  85.2   &    95.4 \\
 & CodeGeeX & - &  - &  22.89  &  39.57   &    60.92 \\[-2pt]
\midrule\\[-10pt]
C++    & \gpt{} & 1   &         tournament &     3.4 &    \textbf{ 55.9} &      \textbf{78.5} \\
       &        & 2   &         tournament &     4.7 &     36.6 &      69.6 \\
       &        & 2   &           lexicase &     4.3 &     35.9 &      73.6 \\
       &        & 4   &         tournament &     4.5 &     27.7 &      62.9 \\
       &        & 10  &         tournament &     3.6 &     13.7 &      49.7 \\
       &        & 16  &         tournament &     4.5 &     13.2 &      51.1 \\
       &        & 100 &         tournament &     3.6 &      7.1 &      31.3 \\[1pt]
\cline{3-7}\\[-8pt]
       & \multicolumn{3}{l}{solved at least once by \method{} with \gpt{}} & 20.1 &     89.0 &      99.4 \\[1pt]
\cline{2-7}\\[-8pt]
    &   Llama 3 & 1   &         tournament &    29.6 &     59.5 &      79.5 \\
       &        & 2   &         tournament &    18.6 &     51.6 &      71.4 \\
       &        & 4   &         tournament &    20.0 &     50.6 &      73.9 \\
       &        & 10  &         tournament &    25.6 &     56.4 &      80.0 \\
       &        & 10  &           lexicase &    25.9 &     \textbf{60.9} &      \textbf{84.2} \\
       &        & 16  &         tournament &    20.9 &     53.0 &      75.5 \\
       &        & 100 &         tournament &    27.1 &     39.7 &      79.2 \\[1pt]
\cline{3-7}\\[-8pt]
       & \multicolumn{3}{l}{solved at least once by \method{} with \llama{}} & 72.0 &     97.6 &      98.8  \\[1pt]
\cline{2-7}\\[-8pt]
& \multicolumn{6}{l}{\textbf{LLM results without \method{}}} \\
& Code Llama 34B & - &  - &  47.8  &  -   &    - \\
& CodeGeeX & - &  - &  17.06  &  32.21   &    51.00 \\[-2pt]
\bottomrule
\end{tabular}
\end{DIFnomarkup} %
\end{adjustbox}
\end{table}

\subsubsection{Overall Generalizability of SEIDR with parent selection strategies and repair-replace trade-off.}
\label{sec:overall-generalizability}

We combine all results in Table~\ref{tab:generalizability-psb2} for PSB2 and Table~\ref{tab:generalizability-he} for HumanEval-X, where we show the number of solved problems at different cutoffs ($k$).  
An average is taken over all runs for the experiments with multiple restarts, and the results for one run are shown for \method{} with Codex. 

Note that at cutoff $k=1$, iterations of \method{} are not taken into account, because the first program candidate is the one LLMs generate from a template, and the trees at cutoff $k=1$ have only one program in each of them, regardless of the parent selection strategy or tree arity. 
Here, the difference between pass@1 results with different parent selection and tree arity is explained by the stochastic behavior of the underlying LLMs. 
We keep pass@1 results to inform the reader about the stochasticity of the models and for the ease of visual comparison of values in each row.

We compare the performance of PSB2 solutions synthesized with \method{} to the PushGP genetic programming system with down-sampled lexicase selection~\cite{helmuth2022:problemsolving}. 
To compare the performance of \method{} with other LLMs without \method{}, we report the average percentage of solved problems for HumanEval-X. 
For HumanEval-X, we cite the performance of the state-of-the-art models as reported by their authors without \method{}, such as CodeGeex~\cite{zheng2023:codegeex}, GPT-3.5 and GPT-4~\cite{openai2023:gpt4}, for reference, and compare \method{} with GPT-3.5 and Code Llama~\cite{roziere2023:code} results at different cutoffs ($k$ for $pass@k$).
Furthermore, we report the number of solved problems in the union of all experiments for a given dataset, model, and language, because different problems are solved in different restarts of the method. 
We refer to these results as ``solved at least once'' hereafter.

In PSB2-Python experiments, \method{} with Codex has a higher pass@1000 than the maximum pass@100 measured for \method{} with \gpt{}. \method{} with Codex also solves 20 problems at least once, whereas \method{} with \gpt{} solves 18.
\method{} with Codex also outperforms other LLMs on PSB2 in C++.
No parent selection strategy or tree arity $N^*$ is leading across all PSB2 experiments. 

In the HumanEval-C++ experiments, \method{} performs well at larger $k,$ i.e., when debugging steps are made. 
Remarkably, the union of \method{} experiments with both \gpt{} and a much smaller \llama{} solve 163 problems (or 99.4\% in Table~\ref{tab:generalizability-he}) and 162 problems (or 98.8\%) in HumanEval-C++, correspondingly.
\method{} does not outperform other LLMs on HumanEval-Python, which, as mentioned earlier, could be the effect of HumanEval-Python being a popular benchmark for testing LLMs and indirectly optimizing the models.

Direct comparison of \method{} results with other iterative program synthesis frameworks is challenging because, in addition to differences in benchmarks used, \citet{jiang2023:selfevolve} and \cite{chen2023:teaching} do not always report the number of generated programs, which is the primary metric that we use to compare iteration against a replace-only baseline.\footnote{\cite{chen2023:teaching} does have a sentence "Note that typically one debugging turn is sufficient"}

\begin{framed}\noindent
\textbf{Summary of \method{} results:} 
\method{} with Codex benefits from the repair-replace trade-off when building a search tree of PSB2 solutions. The best results on PSB2 are achieved with Codex, tree arity of 10 and a maximum of 1000 programs generated. A trend of having the best tree arity set to 10 does not fully hold for other tested models, \gpt{} and \llama{}, where different tree arities performed better in different settings, e.g., depending on the parent selection algorithm, dataset, and programming language. 
Due to increasing costs of newer models time required for testing, we have run \gpt{} and \llama{} six times with each set of hyperparameters and stopped building the search tree at 100 programs. 
In HumanEval-C++, the union of these runs has solved 163 problems with \gpt{}, 162 problems with a much smaller \llama{}-8B model.
\method{} runs have also solved 18 PSB2 problems in C++ and 20 in Python with Codex at least once in all the experiments. 
The numbers for \method{} with \gpt{} are lower than for Codex, possibly due to the focus of \gpt{} on general reasoning and of Codex --- on coding.
A smaller \llama{} model performs poorly on PSB2, which we appoint to the difficulty of the benchmark and the popularity of HumanEval and to the fact that the models can be indirectly optimized for higher performance on HumanEval than on other programming benchmarks.
\end{framed}

\subsection{Threats to Validity}
\label{sec:threats}

External threats to validity concern \method{} performance with benchmarks and language models different from those tested. 
Specifically, PSB2 and HumanEval-X contain programming tasks that require smaller functions to be generated than production-scale software.
Although some canonical solutions in HumanEval-Python have been criticised~\cite{liu2023:your}, we primarily use unit tests to evaluate the output of \method{} and do not compute the exact match. Therefore, these weaknesses do not impact our results.

Internal threats relate to the implementation.
We use PSB2, which has both corner cases and regular tests available. 
To ensure a fair comparison with other studies on PSB2, we evaluate and report results on the provided test set of PSB2, of which we randomly pick 2000 tests. 
A risk that a program does not pass tests other than the ones picked persists but is assumed to be low given the large enough number of 2000 tests. 

The large language models for code editing and text completion used in this study are non-deterministic, 
which can affect the results. 
Due to prohibitive model inference costs and discontinuation of support for earlier models, each experiment with Codex and GPT-3 is run only once. 
Experiments with \llama{} and \gpt{} are restarted six times. 
We acknowledge that more restarts can provide a more complete picture of the approach performance,
although this is not the standard practice in the LLM domain~\cite{ouyang2023:llm} 
and can come at considerable financial and environmental costs.
Our temperature-based sampling procedure described in Section\ref{sec:synth} reduces this stochasticity significantly, especially for low-EPG results, because earlier solutions in the search tree are obtained with lower temperatures, and lower temperature limits the stochasticity.
Codex, GPT-3, and GPT-3.5 are black-box models and can generate malicious code~\cite{pearce2022:asleep}. 
Therefore, when coding in real life, the LLM output should be carefully reviewed before running it.

The results can be skewed towards high performance in the programming languages prevailing in the pre-training dataset used by the authors of the tested LLMs -- an issue also known as \emph{data contamination}.
However, if the models were able to directly reproduce the solutions seen in the training data, 
these problems would have been solved from the first attempt.
Since we observe $EPG>1$ values, data contamination is likely to have only a partial effect on the results.
Moreover, the results can be affected by popular evaluation benchmarks: even though the benchmarks themselves are usually not parts of the pre-training dataset, the published LLMs are likely to be optimized for high performance on these benchmarks.   

\section{Conclusion}
\label{sec:conclusion}

In this study, we propose \method{}, a multi-agent framework to solve the challenge of fully autonomous programming. 
In SEIDR, the program synthesis procedure is augmented from the direct generation of code with large language models instructions to iterative calls to a \debug{} agent followed by the \rank{} agent. 
The \debug{} agent performs a tree search across program candidates generated by a large language model for code.
The LLM used for code repair takes imperfect program candidates and instructions for their improvement as prompts. 
The instructions are obtained from both static templates with failing test case descriptions and templates with auto-generated bug summaries by a text completion language model. 

In addition to the initial exploration of hyperparameters that influence the population size at each generation and the number of children for each parent in \method{} iterations, we extend the framework to test two different ranking strategies (tournament selection and lexicase selection), three models in the coding part of \method{} (Codex, \gpt{} and \llama{}), two datasets (PSB2 and HumanEval-X), two programming languages (Python and C++), and various branching factors. 
With the update to newer models, the prompts and parameters of the tree search are updated accordingly. 

\head{Contributions}
We run one set of initial exploration experiments and two sets of generalizability experiments. 
In the initial exploration, we test \method{} with Codex-edit as the model for draft program synthesis and debugging in Python and C++ on the PSB2 benchmark. 
In our experiments, \method{} outperforms the PushGP baseline and achieves the state-of-the-art result with 19 solved problems out of 25. 
It requires under 1000 program executions to solve them, in stark contrast to billions\footnote{~A problem is considered ``solved'' by PushGP if at least 1 of 100 runs, each with a limit of 60 million programs, was successful.} of executions in PushGP, making it feasible in the areas with costly testing, such as robotics.
Investigation of the repair-replace trade-off shows that \method{} with tree arity of 10 outperforms both the replace-only strategy and the repair-only approach. 

To study the generalizability of \method{}, we experiment with GPT-3.5 and Llama 3-8B as the models for draft program synthesis, explaining errors in synthesized programs and their debugging in Python and C++ on the PSB2 and HumanEval-X benchmarks. 
Our experiments with lexicase and tournament parent selection in the \rank{} agent do not show consistent improvement with either policy or any tree arity. 
One observation is that program candidates in generations prior to the final solution do not pass any test in the test suite in the vast majority of cases. 
The test pass rate for the final solution abruptly increases in one generation from 0 to all tests passed. 

In our generalizability experiments, \method{} shows better performance than using LLMs without \method{} with the same prompts and the same budget in terms of programs to generate.
The method achieves high results of 78.5\% average pass@100 on HumanEval-C++ with \gpt{} and 84.2\% with \llama{}.
Remarkably, in HumanEval-C++, the union of \method{} restarts with different hyperparameters has solved 163 problems with \gpt{}, 162 problems with a much smaller \llama{}-8B model.
The union of \method{} runs has solved 18 PSB2 problems in C++ and 20 in Python with Codex. 

\head{Future work}
Further investigation of \method{} generalizability and ranking strategies are some of the areas for future work. 
Benchmarks with more tests than in HumanEval-X may shed more light on the most effective choice of the number of programs generated from each bug explanation, as well as the framework's comparison on large-context projects. 
As an agent-based framework, \method{} shows how LLM-based agents and other non-LLM agents or components can collaboratively interact and solve software engineering tasks.

\section*{Data Availability}
To support open science and allow for replication and verification of our work, a replication package containing the code and results is publicly available in Zenodo.\footnote{~Zenodo DOI: \href{https://doi.org/10.5281/zenodo.13754705}{10.5281/zenodo.13754705}.
}
\begin{acks}
The work presented in this paper was supported by the European Commission through Horizon 2020 grant 812882, 
and by the Research Council of Norway through the secureIT project (\#288787).
The empirical evaluation made use of the Experimental Infrastructure for Exploration of Exascale Computing (eX3), 
supported by the Research Council of Norway through grant \#270053.
\end{acks}

\bibliographystyle{ACM-Reference-Format}

 %

\end{document}